\documentclass[sigplan,nonacm]{acmart}

\renewcommand\footnotetextcopyrightpermission[1]{}
\setcopyright{none}
\usepackage{tikz}
\usetikzlibrary{arrows.meta}
\usepackage{amsmath}
\usepackage{graphicx}
\usepackage{float}
\usepackage{booktabs}
\usepackage{multirow}
\usepackage{subcaption}
\usepackage{tabularx}
\usepackage{tabulary}
\usepackage{mathtools}
\usepackage{xcolor}
\usepackage{xspace}
\usepackage{listings}
\usepackage{fancyvrb}
\usepackage{fvextra}
\usepackage{pifont}
\input{pyg-perldoc-style}

\newcommand{\bt}[1]{\textcolor{red}{#1 --Baotong}}

\usepackage{titlesec}
\titlespacing*{\section}{0pt}{0.4\baselineskip}{0.2\baselineskip}
\titlespacing*{\subsection}{0pt}{0.3\baselineskip}{0.15\baselineskip}

\usepackage[font=small]{caption}

\usepackage{enumitem}
\setlist[itemize]{noitemsep, topsep=0pt, leftmargin=1em}

\usepackage{tikz}
\newcommand*\circled[1]{\tikz[baseline=(char.base)]{
            \node[shape=circle,fill=.,inner sep=0pt] (char) {\color{-.}\textsf\footnotesize #1};}}

\newcommand{\minisec}[1]{\vspace{0.1em}\noindent\textbf{#1. }}
\newcommand{\myparagraph}[1]{\noindent\textbf{#1. }}

\newenvironment{packed_itemize}{
    \begin{list}{\labelitemi}{\leftmargin=1.0em}
     \setlength{\itemsep}{2.5pt}
     \setlength{\parskip}{0pt}
     \setlength{\parsep}{0pt}
     \setlength{\headsep}{0pt}
     \setlength{\topskip}{0pt}
     \setlength{\topmargin}{0pt}
     \setlength{\topsep}{0pt}
     \setlength{\partopsep}{0pt}
    }{\end{list}}

\newcommand{\sysname}{\textsc{Spin}\xspace}

\definecolor{codehighlight}{HTML}{FFF0B3}
\definecolor{codegreen}{rgb}{0,0.6,0}
\definecolor{codegray}{rgb}{0.5,0.5,0.5}
\definecolor{codepurple}{rgb}{0.58,0,0.82}
\definecolor{backcolour}{rgb}{0.95,0.95,0.92}

\lstdefinestyle{pycode}{
    backgroundcolor=\color{backcolour},   
    commentstyle=\color{codegreen},
    keywordstyle=\color{magenta},
    numberstyle=\tiny\color{codegray},
    stringstyle=\color{codepurple},
    basicstyle=\ttfamily\footnotesize,
    breakatwhitespace=false,         
    breaklines=true,                 
    captionpos=b,                    
    keepspaces=true,                 
    numbers=left,                    
    numbersep=5pt,                  
    showspaces=false,                
    showstringspaces=false,
    showtabs=false,                  
    tabsize=2
}
\begin{document}


\title[]{Unifying Sparse Attention with Hierarchical Memory for Scalable Long-Context LLM Serving}
\author{Zihan Zhao}
\authornote{Work performed during the internship while at Microsoft Research.}
\affiliation{%
  \institution{University of Virginia}
  \country{}}
\email{rxy6cc@virginia.edu}

\author{Baotong Lu}
\affiliation{%
  \institution{Microsoft Research}
  \country{}}
\email{baotonglu@microsoft.com}

\author{Shengjie Lin}
\authornotemark[1]
\affiliation{%
  \institution{Georgia Institute of Technology}
  \country{}}
\email{slin468@gatech.edu}

\author{Yizou Chen}
\affiliation{%
  \institution{The Chinese University of Hong Kong}
  \country{}}
\email{chenyz@cse.cuhk.edu.hk}

\author{Jing Liu}
\affiliation{%
  \institution{Microsoft Research}
  \country{}}
\email{jingliu3@microsoft.com}

\author{Yanqi Zhang}
\affiliation{%
  \institution{Microsoft Research}
  \country{}}
\email{yanqizhang@microsoft.com}

\author{Ziming Miao}
\affiliation{%
  \institution{Microsoft Research}
  \country{}}
\email{Ziming.Miao@microsoft.com}

\author{Ming-Chang Yang}
\affiliation{%
  \institution{The Chinese University of Hong Kong}
  \country{}}
\email{mcyang@cse.cuhk.edu.hk}

\author{Haiying Shen}
\affiliation{%
  \institution{University of Virginia}
  \country{}}
\email{hs6ms@virginia.edu}

\author{Qi Chen}
\affiliation{%
  \institution{Microsoft Research}
  \country{}}
\email{cheqi@microsoft.com}

\author{Fan Yang}
\affiliation{%
  \institution{Microsoft Research}
  \country{}}
\email{fanyang@microsoft.com}

\renewcommand{\shortauthors}{Zhao et al.}

\begin{abstract}

Long-context LLM serving is bottlenecked by the cost of attending over ever-growing KV caches. Dynamic sparse attention promises relief by accessing only a small, query-dependent subset of the KV state per decoding step and extending the KV storage to CPU memory.
In practice, however, these algorithmic savings rarely translate into end-to-end system-level gains because sparse methods typically operate at different granularities and thus rely
on ad hoc, per-algorithm implementations.
At the same time, hierarchical KV storage introduces a new systems bottleneck: retrieving fine-grained, irregular KV subsets across the GPU--CPU boundary can easily erase the benefits of sparsity.

We present \sysname, a sparse-attention-aware inference framework that co-designs the execution pipeline with hierarchical KV storage through three techniques: (1) a unified \emph{partition} abstraction that maps different sparsity granularities onto a shared page-based KV substrate; (2) a locality-aware KV cache manager that dynamically sizes per-request HBM budgets and uses a GPU-friendly bucketed LRU policy to cut PCIe round-trips; and (3) a two-level hierarchical metadata layout sized to the active working set rather than the worst-case address space. Built on vLLM with three representative sparse attention algorithms, \sysname delivers $1.66$--$5.66\times$ higher end-to-end throughput and $7$--$9\times$ lower TTFT than vLLM, and reduces TPOT by up to $58\%$ over the original sparse-attention implementations.

\end{abstract}

\maketitle


\section{Introduction}
\label{sec:intro}

To support increasingly sophisticated workloads such as long-context reasoning~\cite{wei2022chain, bai-etal-2024-longbench, geminiteam2024gemini15unlockingmultimodal}, document understanding~\cite{jin2024long, wang2023docllmlayoutawaregenerativelanguage, ma2024mmlongbenchdocbenchmarkinglongcontextdocument}, and code generation\cite{bairi2024codeplan, roziere2024codellamaopenfoundation, zhang-etal-2023-repocoder}, modern LLMs have expanded their context windows to hundreds of thousands or even millions of tokens\cite{chatgpt, claude, gemini, llama4}. 
This trend makes long-context serving fundamentally costly because the key-value (KV) cache~\cite{pope2023efficiently}, intermediate states in \textit{attention}, grows linearly with the sequence length, and decoding repeatedly reads all historical KV states for every generated token.
Consequently, serving is jointly constrained by 
the memory bandwidth to sustain linear KV accesses and 
GPU memory capacity to hold expanding KV caches.

Dynamic sparse attention offers a promising algorithmic response. Prior work shows that, for a given query, only a small subset of historical tokens typically dominates the next-token prediction~\cite{child2019generating,correia2019adaptively,deng2024sparse}. Recent sparse methods therefore preserve the full KV cache but select only the critical subset at each decoding step\cite{tangQUESTQueryAwareSparsity2024,liu2024retrievalattention,magicpig,yuan2025native}, reducing the amount of KV data that attention must process per step.

The selective access pattern of sparse attention breaks the full-KV dependency assumed by prior dense-serving systems~\cite{kwonEfficientMemoryManagement2023,zheng2024sglang} and makes hierarchical KV storage attractive. 
The system can keep the full KV cache in high-capacity CPU memory and fetch only critical KV states to the GPU on demand at fine granularity~\cite{chenRetroInferVectorStorageApproach2025,sunShadowKVKVCache2025a}, thereby alleviating     GPU memory pressure.
For example, a single 128K-token request for Qwen3-8B requires roughly 19GB of KV cache, whereas a sparse method with 95\% sparsity needs to process less than 1GB of KV data per decoding step, enabling higher concurrency and better hardware utilization.

Yet, these algorithmic gains do not automatically translate into system-level efficiency, for two major challenges.

\begin{figure}[t]
    \centering
    \includegraphics[width=\columnwidth]{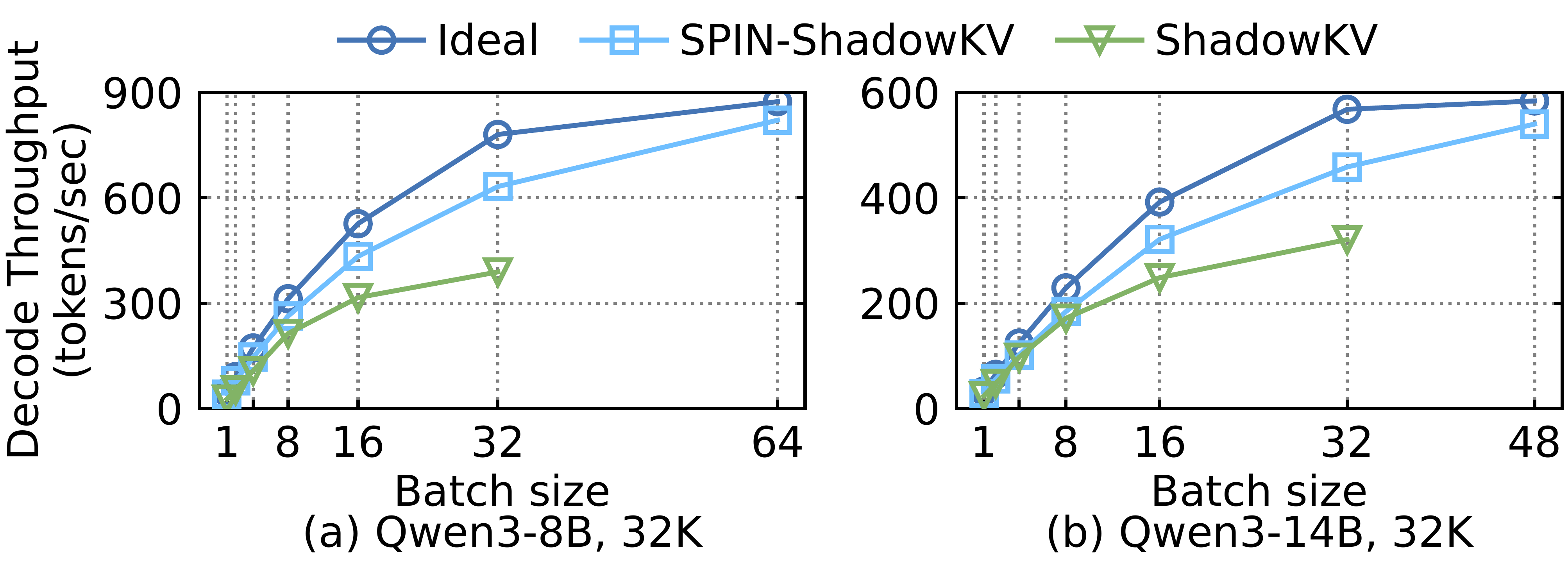}
    \caption{
    {\textbf{Gap Between Theoretical and Realized Performance.}} 
    The ideal envelope represents an oracle system with optimal GPU--CPU KV caching and retrieval.
    On 32K-context workloads on an A100 GPU (80GB), existing sparse attention implementations~\cite{sunShadowKVKVCache2025a} remain far below this bound and also fail to scale to large batch sizes due to inefficient memory management, whereas the \sysname{}-powered system approaches the envelope.
    }
    \label{fig:intro}
\end{figure}

\myparagraph{Lack of Unified Abstractions for Sparsity} 
Sparse algorithms are diverse, rapidly evolving, and operate at different granularities (e.g., blocks~\cite{sunShadowKVKVCache2025a} and clusters~\cite{chenRetroInferVectorStorageApproach2025}). Existing inference frameworks~\cite{kwon2023efficient,zheng2024sglang} do not provide a common interface that accommodates these variants, so existing sparse-attention support has evolved into algorithm-specific prototypes that cannot readily share optimized kernels, memory management, or online scheduling support. 
Realizing the system-level benefits of sparsity therefore requires a unified abstraction that preserves algorithm-specific flexibility while enabling the system to optimize the common substrate.

\myparagraph{Irregular KV Retrieval Erodes Sparsity Benefits} 
Once the full KV cache is placed in CPU memory, sparse serving becomes constrained by GPU--CPU data movement. Unlike dense attention, whose KV accesses are largely contiguous and predictable\cite{qin2025mooncake}, sparse attention retrieves small, scattered KV subsets that vary across decoding steps, layers, and heads. These irregular transfers underutilize PCIe bandwidth and add latency to the critical path. Prior work~\cite{tokenselect, xu2024recycled} shows that autoregressive decoding also exhibits temporal locality: critical tokens often overlap across consecutive steps. Realizing the benefits of sparsity therefore fundamentally requires a locality-aware memory manager that both caches useful KV states and retrieves misses efficiently.

To better quantify the gap between the theoretical and realized performance of sparse attention algorithms, we define an \textit{ideal sparse-serving envelope}: an oracle system that serves most selected KV tokens from GPU memory while incurring the 
theoretically minimum retrieval overhead under an optimal caching policy~\cite{belady1966study}. As shown in Figure~\ref{fig:intro}, existing sparse attention implementations remain far below this bound due to the aforementioned reasons.

To narrow the gap between sparse attention's theoretical promise and practical system efficiency, we present \sysname{}, a \underline{sp}arse-attention-aware \underline{in}ference framework. 
\sysname abstracts diverse sparse attention algorithms into a common five-operation pipeline with three algorithm-specific compute operations (\textit{Index}, \textit{Select}, and \textit{Attention}) and two shared data-management operations (\textit{Offload} and \textit{Retrieve}). The key observation is that these algorithms share a common execution pattern and differ mainly in sparsity granularity. \sysname therefore introduces \textit{partition} as a common logical unit that maps different granularities onto the same pipeline. This design exposes clean interfaces for algorithm-specific customization while reusing optimized implementations of \textit{Offload} and \textit{Retrieve}, 
 allowing diverse sparse methods to share efficient KV management.

Based on this pipeline, \sysname builds a KV-cache management system for sparse serving across the GPU--CPU hierarchy with three design decisions:

\myparagraph{1. Partition Abstraction}
\sysname{} treats partition as the
key abstraction connecting sparse algorithms with page-based
KV cache management. This separates algorithm-defined
sparsity units from hardware-efficient units, allowing the runtime to preserve algorithmic flexibility while
allocating and transferring KV data at page granularity.

\myparagraph{2. Dynamic, Locality-aware KV Management}
To maximize GPU residency under dynamic serving workloads, \sysname{}
extends beyond conventional append-only paged KV caches with
a lightweight, locality-aware KV manager. Its dynamic KV
buffer supports per-request resizing and in-place updates as
requests arrive, advance, and complete, while maintaining
GPU-friendly metadata that enables efficient locality-aware
replacement policies.

\myparagraph{3. Hierarchical Metadata}
Metadata overhead such as page tables increases with
context length: even dense attention incurs nontrivial
bookkeeping, and attention head-wise sparsity pattern introduces additional indexing metadata. \sysname therefore adopts
OS-style multi-level indexing so that metadata scales with
the physical working set rather than the worst-case logical
space, while splitting metadata across GPU and CPU to reduce on-GPU overhead.

We implement \sysname on top of vLLM and integrate three representative sparse attention algorithms~\cite{sunShadowKVKVCache2025a,chenRetroInferVectorStorageApproach2025,gao2025seerattention-r}, demonstrating the generality of the framework. 
We evaluate \sysname on three widely used models~\cite{qwen3, llama3.1-70B} spanning different scales, two long-context benchmarks~\cite{longbench-v2, liu2024longgenbench}, and both A100 and B200 GPUs.
Across models, hardware platforms, and workloads, \sysname delivers $1.66$--$5.66\times$ higher end-to-end throughput, $7$--$9 \times$ lower TTFT than vLLM, with larger gains at higher request rates.
\sysname also improves throughput by up to $2.39\times$ over the original implementations of sparse attention, demonstrating the effectiveness of system-level optimizations.

\section{Background}
\label{sec:bg}


\subsection{Long-Context LLM Serving}

\myparagraph{Challenges of Long-Context Serving}
The context window of state-of-the-art LLMs and emerging workloads continues to grow, stressing inference systems along two dimensions~\cite{wei2022chain,jin2024long,bairi2024codeplan}.
First, 
since KV cache grows linearly with sequence length, 
it creates severe GPU memory-capacity pressure, reducing the number of requests that can remain resident on the device.
Scaling to more GPUs can expand effective capacity, but at a higher provisioning cost. 
Many systems therefore offload KV states to larger CPU memory and keep only an active subset on the GPU~\cite{chenRetroInferVectorStorageApproach2025,sunShadowKVKVCache2025a,lee2024InfiniGen,sheng2023flexgen}. 

Second, decoding becomes memory-bandwidth-bound because each step must access all historical KV states. Even when the full cache fits in HBM, the decoding speed is limited by GPU memory bandwidth~\cite{chen2023accelerating,ribarsparq}. Once the KV cache is partially offloaded, PCIe transfers also enter the critical path and can dominate latency.

\subsection{Dynamic Sparse Attention}
A key opportunity is that attention is inherently sparse in practice. 
Sparse attention~\cite{h2o,tangQUESTQueryAwareSparsity2024,sunShadowKVKVCache2025a, chenArkValeEfficientGenerative2024} exploits this property by computing attention over only important tokens instead of the full context while preserving model accuracy.
Leveraging sparsity reduces GPU computation, lowers KV accesses, and alleviates bandwidth pressure for long-context inference. It also makes hierarchical KV storage more attractive: the full cache can remain in high-capacity CPU memory while the system fetches only critical tokens to the GPU. 
The challenge, however, is that token importance dynamically changes across decoding steps, model layers, and attention heads, so the sparse attention mechanism must identify critical tokens accurately and efficiently.

\myparagraph{Variety of Sparse Attention Algorithms}
Given the growing interest in sparse attention, many variants have been proposed~\cite{child2019generating,correia2019adaptively,deng2024sparse,tangQUESTQueryAwareSparsity2024,liu2024retrievalattention,magicpig,chenRetroInferVectorStorageApproach2025,yuan2025native, liu2025deepseek}. To preserve accuracy, modern algorithms typically maintain the full KV cache and dynamically select a subset of tokens across decoding steps. Their main difference is how they organize the KVs and identify critical ones, with different trade-offs.
For example,
ShadowKV~\cite{sunShadowKVKVCache2025a} partition the KV cache into fixed-size blocks and 
compute a summary (i.e., average key) for each block such that 
the query can compute with summaries to select promising blocks~\cite{tangQUESTQueryAwareSparsity2024,sunShadowKVKVCache2025a}; 
RetroInfer~\cite{chenRetroInferVectorStorageApproach2025} instead groups similar key vectors into clusters for more accurate selection.

\myparagraph{Common Design Pattern}
Across different algorithms, the common pattern is to partition the context into a coarser logical unit and use an auxiliary sketch (e.g., summaries) to select which parts of the KV cache should participate in attention. The design space therefore centers on the partition granularity and on how efficiently the system can build, query, and retrieve from this structure at decoding.

\myparagraph{Need for Runtime Support}
Instead of sequentially and repeatedly reading a uniform KV cache, sparse attention must execute a selection-and-retrieval pipeline whose accesses are scattered and vary a lot. Dense-serving runtimes~\cite{kwonEfficientMemoryManagement2023} are a poor fit for this pattern because their KV managers, kernels, and schedulers are designed for full-cache access. 
As a result, existing sparse attention implementations are typically algorithm-specific, ad-hoc prototypes~\cite{sunShadowKVKVCache2025a,chenRetroInferVectorStorageApproach2025}: each method re-implements its own KV-cache management and inference pipeline, often without mature runtime components such as scheduling, which raises porting effort and limits reuse of optimized system support. 

\section{Towards Ideal Sparsity Serving}
\label{sec:principle}

Dynamic sparse attention reduces the amount of KV accesses, but it does not reduce the total KV cache footprint. As a result, practical sparse-serving systems typically rely on a hierarchical KV store, with the full cache placed in CPU memory and only a working set kept on the GPU. 
Under this setting, the benefit of sparsity is no longer determined by attention sparsity alone, but by how effectively the runtime orchestrates data movement across the GPU--CPU boundary to match the dynamic access patterns induced by sparsity.

This observation shifts the core question of sparse serving from sparse computation to sparse data management. To realize the algorithmic promise of sparsity, the serving system must jointly optimize two factors: GPU KV-cache residency and GPU-CPU data movement efficiency. 
If either is poorly handled, sparse attention can deliver little practical gain and may even underperform highly optimized dense-serving systems, far from its theoretical potential under optimal KV caching and PCIe utilization.

\subsection{The Promise vs. Reality of Sparse Serving}

To quantify the theoretical upper bound, we construct an \emph{ideal sparse-serving envelope}---a hypothetical system that simultaneously achieves optimal KV caching and perfect data movement.
Let $B$ denote the batch size, $L$ the number of layers, $H$ the number of KV heads, $d$ the head dimension, $e$ bytes per element, and $N$ the context length. A sparse attention algorithm is characterized by two ratios: $\alpha$, the size of the compressed summary vectors (used for fast importance scoring) relative to the full KV cache; and $\beta$, the fraction of the full KV cache selected for attention computation. At the system level, $\rho$ denotes the GPU cache miss ratio---the fraction of selected KV data that is not resident on the GPU and must be fetched from CPU memory. The HBM data volumes for scoring and sparse attention are thus modeled as follows.
\begin{equation}
  QK_{\text{score}}
    = B \cdot L \cdot H \cdot d \cdot e \cdot \alpha N
    \label{eq:vol_sel}
\end{equation}
\vspace{-8pt}
\begin{equation}
  KV_{\text{topk}}
    = 2 \cdot B \cdot L \cdot H \cdot d \cdot e \cdot \beta N
    \label{eq:vol_attn}
\end{equation}
Among the $KV_{\text{topk}}$ bytes needed for attention, a fraction $\rho$ must be fetched from CPU memory over PCIe; the remainder is served from HBM. Let $\mathcal{B}_{HBM}$ and $\mathcal{B}_{PCIe}$ denote the effective HBM and PCIe bandwidth, and $T_{\text{MLP}}$ the MLP latency per step. The time per output token (TPOT) is defined by
\begin{equation}
  \text{TPOT} =
    \frac{QK_{\text{score}} + KV_{\text{topk}}}{\mathcal{B}_{HBM}}
    +
    \frac{\rho \cdot KV_{\text{topk}}}{\mathcal{B}_{PCIe}}
    + T_{\text{MLP}}
    \label{eq:tpot}
\end{equation}
\vspace{-4pt}

The ideal curve is obtained by setting each term in Equation~\ref{eq:tpot} to its theoretical minimum: (1)~Belady-optimal eviction~\cite{belady1966study} to minimize $\rho$ given the available GPU cache capacity; (2)~peak $\mathcal{B}_{PCIe}$ utilization for retrieving cache misses; and (3)~peak $\mathcal{B}_{HBM}$ utilization for all GPU-resident accesses.

However, such a system does not exist in practice. First, optimal caching is unattainable because it requires perfect foresight of future token accesses, which cannot be predicted exactly from the model's evolving attention patterns. Second, achieving full PCIe utilization is fundamentally difficult because the KV tokens retrieved from CPU memory are often sparse, strided, and non-contiguous, as dictated by the sparse algorithms. As a result, the gap between this idealized optimum and practical systems is inevitable, and is precisely where the systems challenge of sparse serving lies. 

\begin{figure}[t]
    \centering
    \includegraphics[width=\linewidth]{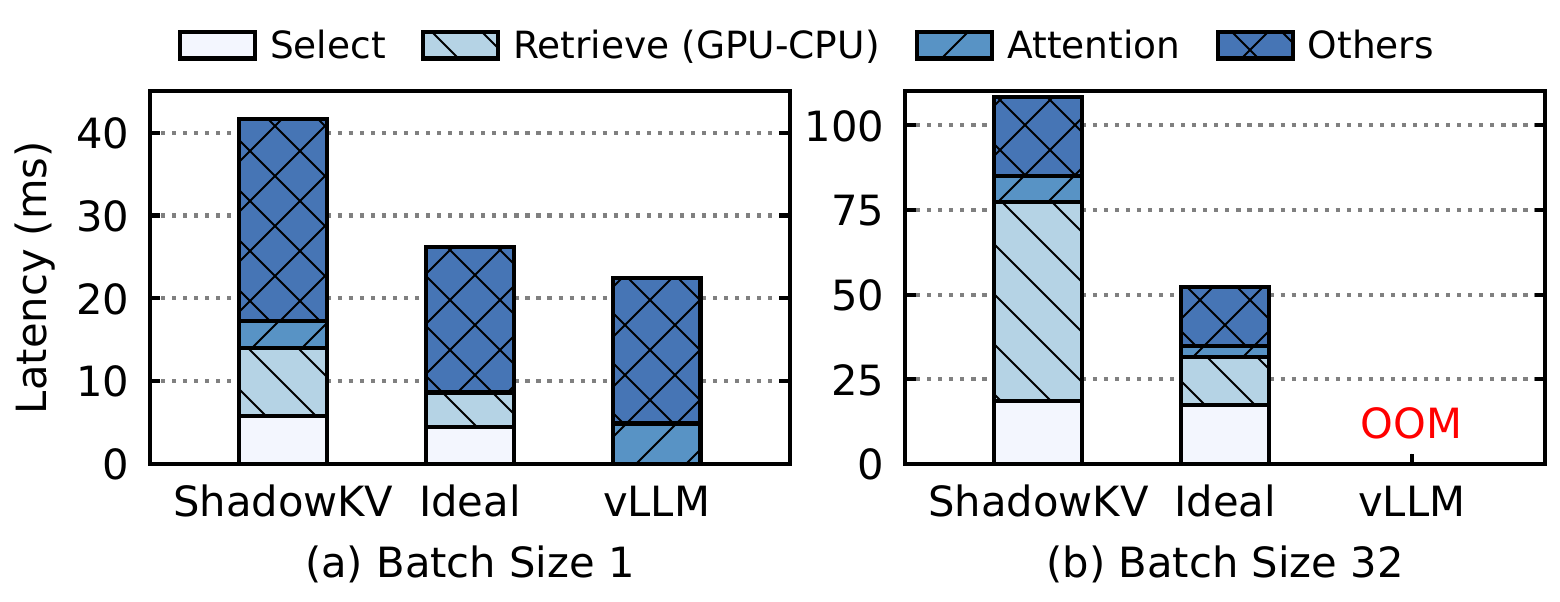}
    \caption{{\textbf{Per-token Decode Latency Breakdown of Different Systems.}}
    Existing sparse attention implementation~\cite{sunShadowKVKVCache2025a} incurs much higher decode latency than the ideal sparse-serving envelope, and significantly underperforms a dense-serving system (vLLM~\cite{kwon2023efficient}) at low load. 
    }
    \label{fig:mot_breakdown}
\end{figure}

To illustrate this gap, we compare the per-token decode latency of an existing sparse attention implementation (ShadowKV~\cite{sunShadowKVKVCache2025a}) against the ideal sparse-serving simulator and a dense-serving system (vLLM~\cite{kwon2023efficient}) across different batch sizes\footnote{Evaluated on Qwen3-14B with 32K-context workloads on a A100 GPU.}.
The results, shown in Figure~\ref{fig:mot_breakdown}, reveal a significant gap between ShadowKV and the ideal curve, with ShadowKV incurring much higher latency than the ideal system. At low batch sizes, the ideal curve underperforms vLLM because attention is not yet a bottleneck while the introduced scoring and retrieval overhead are not negligible, but ShadowKV still incurs 2$\times$ retrieval overhead than the ideal curve. As batch size increases, the gap to the ideal curve further widens due to higher data movement overhead over PCIe.

\subsection{From Promise to Reality: The System Challenges}

Narraowing the gap between the promise and reality of sparse serving requires addressing three system challenges:

    \myparagraph{General System Abstraction for Sparse Attention}
      Sparse attention algorithms span a wide spectrum of granularities, 
      such as blocks or clusters with varying sizes.
      A general sparse-serving system must therefore provide a common inference-pipeline abstraction that makes different algorithms easy to integrate while hiding granularity differences behind a unified interface. Moreover, this abstraction must preserve common optimizations across algorithms.

    \myparagraph{Efficient GPU--CPU KV Cache Management}
    To minimize GPU--CPU data movement, the system must maximize the GPU residency of critical KV vectors. This requires a locality-aware buffer manager that retains reusable partitions using efficient cache replacement, while maintaining high performance under dynamic serving workloads. When cache misses are unavoidable, it must also retrieve fine-grained, non-contiguous KV pages efficiently over PCIe.

    \myparagraph{Scalable Metadata Management}
    Head-wise sparsity requires per-head page tables to track KV residency across GPU and CPU and to drive cache replacement. 
    At long context lengths, these metadata can become substantial, potentially reaching tens or hundreds of gigabytes. 
    The root cause is that existing serving systems such as vLLM typically pre-allocate metadata arrays for a worst-case logical configuration; under head-wise sparse serving, structures such as page tables therefore scale with the maximum context length and the number of heads rather than the active working set. 
    Metadata management must therefore keep this bookkeeping compact, especially on the GPU, so that metadata does not crowd out effective KV-cache capacity. 

    

\section{\sysname Interfaces}
\label{sec:sys_interface}

\begin{figure}[t]
    \centering
    \includegraphics[width=\linewidth]{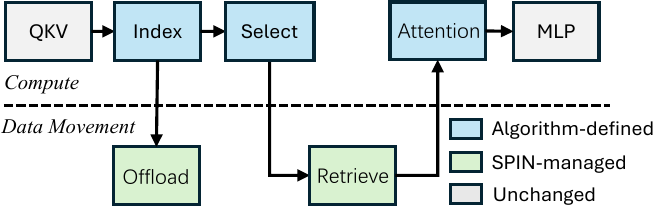}
    \caption{The abstracted inference pipeline for sparse attention in \sysname, illustrating the decoupling of computation and data movement. Select and Retrieve are only invoked during the decode.}
    \label{fig:inf_pipe}
\end{figure}

\begin{table*}[t]
    \footnotesize
    \centering
    \caption{\textbf{Operation mappings across various sparse attention algorithms.}  SeerAttention-R uses head-wise linear projection for pooled query and key, i.e., \texttt{`bshi,hio->bsho'}, denoted by \texttt{exp}. The Attention operator is omitted as it largely follows standard FlashAttention~\cite{dao2022flashattention}. The copy kernel used in Offload and Retrieve moves data from source region (1st parameter) to destination region (2nd parameter).}
    \renewcommand{\arraystretch}{1.2}
    \newcolumntype{C}{>{\centering\arraybackslash}X}
    \begin{tabularx}{\textwidth}{l | c | C | c | c}
        \toprule
         & \textbf{Index} & \textbf{Select} & \textbf{Offload} & \textbf{Retrieve} \\
        \midrule
        \textbf{ShadowKV} & \texttt{mean(paged\_key, dim=dim\_page)} & \multirow{2}{*}{\texttt{score(roped\_q, k\_sums) $\rightarrow$ topk}} & \multirow{4}{*}{\begin{tabular}[c]{@{}l@{}}\texttt{copy(gpu\_kv,}\\ \phantom{\texttt{copy(}}\texttt{cpu\_kv,}\\ \phantom{\texttt{copy(}}\texttt{map)}\end{tabular}} & \multirow{4}{*}{\begin{tabular}[c]{@{}l@{}}\texttt{copy(cpu\_kv,}\\ \phantom{\texttt{copy(}}\texttt{gpu\_kv,}\\ \phantom{\texttt{copy(}}\texttt{indices)}\end{tabular}} \\
        \cline{1-2}
        \textbf{RetroInfer} & \texttt{segment\_k\_means(main\_key)} & & & \\
        \cline{1-3}
        \textbf{SeerAttention-R} & \texttt{einsum(exp, pooled\_key, W\_kg)} & \begin{tabular}[c]{@{}c@{}}\texttt{einsum(exp, pooled\_q, W\_qg) $\rightarrow$ gated\_q} \\ \texttt{score(gated\_q, k\_sums) $\rightarrow$ topk}\end{tabular} & & \\
        \bottomrule
    \end{tabularx}
    \label{tab:abs_table}
\end{table*}



We introduce \sysname, a sparse-attention-aware inference engine that manages the KV cache across the GPU--CPU memory hierarchy. 
It is designed to support a range of sparse attention algorithms by transparently translating algorithmic sparsity into system-level efficiency.
This section presents \sysname's abstract interfaces for algorithm integration.

\subsection{Sparse Attention Abstraction}

To unify sparse attention algorithms with different sparsity granularities, \sysname uses \textit{partition} as a common logical unit.
A partition is a contiguous group of tokens that serves as the unit of selection and retrieval; it may represent a block, a cluster, or a single token.
Moreover,
despite differences in how algorithms identify critical partitions, they share the same execution pattern: each decoding step attends to only a subset of the full KV cache.
\sysname therefore abstracts this common structure as the five-stage pipeline in Figure~\ref{fig:inf_pipe}, mapping diverse algorithms onto a unified execution flow.

These operations fall into two categories: \textit{algorithm-defined compute} and \textit{\sysname-managed data movement}.
\textit{Index}, \textit{Select}, and \textit{Attention} implement the algorithm-specific logic for constructing partitions, selecting critical partitions, and computing attention.
\textit{Offload} and \textit{Retrieve}, in contrast, manage KV placement and movement across the GPU--CPU memory hierarchy.
This split decouples algorithmic logic from system-level data management, allowing diverse sparse methods to share a common runtime.
Table~\ref{tab:abs_table} summarizes the mapping of representative sparse attention algorithms to these operations.
Specifically, these operations are defined as follows:

{\renewcommand{\labelitemi}{\scriptsize$\bullet$}
\begin{packed_itemize}
\item \textbf{Index}.
Index runs in both prefill and decode to organize KV tokens into partitions.
During prefill, it consumes the prompt KV cache and produces a \textit{partition specification}, per-partition summaries, and other algorithm-specific metadata.
The partition specification defines how tokens are grouped and is passed to Offload to materialize the physical layout.
As shown in Table~\ref{tab:abs_table}, the internal logic of Index varies across algorithms.
During decode, Index is invoked periodically to incorporate newly generated tokens.

\item \textbf{Offload}. 
Given the partition specification and summaries from Index, Offload manages KV placement across GPU and CPU memory.
It keeps summaries in GPU memory for fast selection, materializes the partitioned KV cache asynchronously, places partitions in the appropriate memory tier, and updates \sysname metadata to track their locations.

\item \textbf{Select}.
Select is decode-specific.
According to the current query and the summaries of all partitions, it identifies most critical partitions for the current decoding step.

\item \textbf{Retrieve}.
Given the indices of critical partitions, Retrieve checks their current residence and brings any missing partitions from CPU memory back to GPU memory.

\item \textbf{Attention}. 
Attention computes the output using the selected KV tokens.
For most algorithms, this stage reuses a standard FlashAttention kernel.
Some algorithms, however, require customized attention kernels; for example, RetroInfer~\cite{chenRetroInferVectorStorageApproach2025} incorporates additional estimation logic beyond standard attention.
\sysname supports this by allowing algorithms to register custom attention operators.
\end{packed_itemize}
}

In summary, \sysname exposes Index, Select, and Attention as the customization points for algorithm developers, while reusing optimized system support for Offload and Retrieve.
By managing the KV cache via partitions throughout the pipeline, \sysname hides low-level memory management and enables efficient execution across sparse attention methods.

\begin{figure}[t]
    \centering
    \begin{minipage}{\linewidth}
    \input{pyg-shadowkv.tex}
    \end{minipage}
    
    \caption{Pseudocode illustrations demonstrating the integration of ShadowKV~\cite{sunShadowKVKVCache2025a} into \sysname. Highlighted lines (4--9 and 15--16) denote the core algorithmic logic, directly copied from the original implementation.}
    \label{fig:demo}
\end{figure}

\subsection{Integrating Sparse Attention Algorithms}

Integrating a sparse attention algorithm into \sysname requires implementing only its algorithm-specific logic. \sysname exposes an abstract Python class, \texttt{SparseAttention}, through which developers define \texttt{index} and \texttt{select}, and optionally a custom attention operator, then register them at the appropriate stages of the inference pipeline. For example, ShadowKV and RetroInfer register after RoPE, and SeerAttention-R before RoPE because it operates on pre-RoPE queries and keys.

Figure~\ref{fig:demo} illustrates integration using ShadowKV as an example, with highlights marking the algorithm-specific code that must be ported. In ShadowKV, Index performs mean pooling on the page dimension to generate landmarks as summaries, determines outliers from cosine similarities, filters out the recent window, and assigns the remaining to offloading. The computed page IDs are then used to construct the partition map, where outliers and the recent window are kept in GPU memory and always participate in attention, while the rest are offloaded to CPU memory. Because ShadowKV uses uniform partitioning --- every partition owns exactly one page --- page IDs are sufficient for partition organization. To support non-uniform partitioning like RetroInfer~\cite{chenRetroInferVectorStorageApproach2025}, algorithm-computed partition-to-token mappings are used to construct the partition specification. On the other hand, ShadowKV's Select simply implements top-$k$ scoring.

\section{Memory Management}
\label{sec:mem-mgmt}

\subsection{Overview}

\begin{figure}[t]
    \centering
    \includegraphics[width=\linewidth]{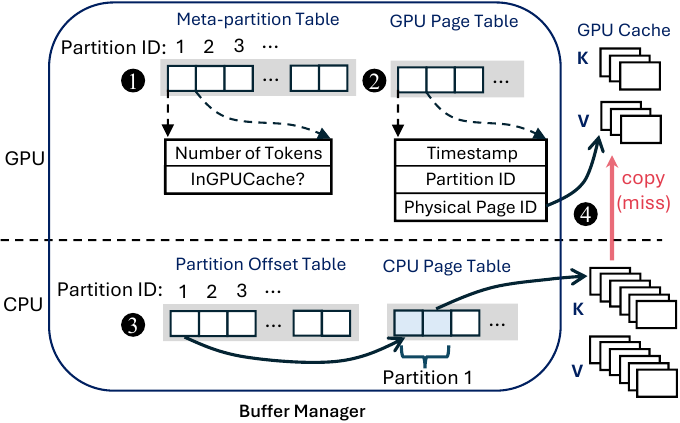}
    \caption{Overview of \sysname's memory management system. 
    Steps 1--4 describe the process of ``Retrieve'' for critical partitions. 
    }
    \label{fig:overview}
\end{figure}

As discussed in \S~\ref{sec:principle}, the practical performance of sparse attention ultimately hinges on a serving engine that can efficiently manage the KV-cache across the GPU--CPU memory hierarchy, thereby minimizing costly PCIe transfers during the Offload and Retrieval operations. In particular, the memory manager must address three foundamental challenges: reconciling the mismatch between algorithmic access granularities and hardware-efficient block accesses, maximizing GPU KV cache residency under dynamic workloads, and minimizing metadata overhead at long context lengths. \sysname addresses these challenges with three corresponding techniques:

\begin{packed_itemize}
\item \textbf{Partition Abstraction}:
\sysname elevates \textit{partition} to the core abstraction between sparse algorithms and page-based KV-cache management. This decouples algorithm-defined sparsity units from hardware-efficient page granularity, allowing \sysname to preserve both flexibility and I/O efficiency.

\item \textbf{Dynamic, Locality-aware KV Management}:
To maximize GPU residency under dynamic serving workloads, \sysname goes beyond conventional append-only paged KV caches and introduces a lightweight, locality-aware KV manager. Its dynamic KV buffer supports per-request resizing and in-place updates as requests arrive, progress, and depart, while maintaining GPU-friendly metadata for efficient locality-aware replacement policies.

\item \textbf{Hierarchical Metadata}:
Metadata scales linearly with context length. As context length increases, dense attention already incurs nontrivial bookkeeping overhead, and sparse attention adds further indexing metadata. To contain this cost, \sysname borrows OS-style multi-level indexing so that metadata scales with the physical working set rather than the worst-case logical space, and splits metadata across GPU and CPU to minimize on-GPU overhead.
\end{packed_itemize}

Combining all three techniques, Figure~\ref{fig:overview} presents the overall architecture of \sysname's memory manager. The partition abstraction is implemented through four metadata tables that are multi-level indexed and split across GPU and CPU memory. In particular, \sysname adopts a page-centric organization on the GPU, maintaining page-to-partition mappings to support efficient KV-cache replacement. This design enables a custom LRU kernel that fully exploits GPU parallelism to select victim on-GPU KV-cache entries and to perform bulk updates when the KV caches of many requests change under dynamic serving workloads. 
Next, we describe each of the techniques in more details.



\subsection{Partition Abstraction}
\label{subsec:partition}

\sysname uses fixed-size pages as the unit of memory allocation and data transfer to simplify memory management, reduce fragmentation, and preserve high I/O efficiency.
Above this physical layer, \textit{partitions} define the logical granularity. A partition may span one or more pages, and we employ the partition-to-page mapping to translate each logical partition ID to the physical page IDs in GPU memory 
or CPU memory. 

\minisec{Head-wise Physical Pages}
\sysname adopts head-wise physical pages because different heads often attend to different critical tokens~\cite{sunShadowKVKVCache2025a,chenRetroInferVectorStorageApproach2025}.
Unlike dense serving layouts that typically colocate KV vectors of all heads in one page, this design avoids fetching unused data over PCIe during retrieval. 

\minisec{Tier-Split Mapping Table}
Supporting variable-length partitions requires a compact representation of page lists for efficient GPU access, while the mapping metadata must be placed carefully to balance access latency against memory overhead. \sysname addresses both requirements with a tier-split mapping that keeps latency-critical metadata on the GPU and places capacity-dominating metadata in CPU memory.

The mapping uses four tables. On the GPU, the Meta-partition table stores each partition's token count and residency state, while the GPU page table tracks cached partitions' GPU-resident page IDs and stores the recency timestamps used for cache replacement. On the CPU, the Partition Offset table maps each partition ID to the starting offset of its page list, and the CPU page table stores the corresponding physical CPU page IDs in a flat contiguous array for efficient GPU access. Together, these CPU-side tables provide a forward mapping from partition IDs to CPU pages. To reduce GPU memory overhead without sacrificing GPU-controlled retrieval, \sysname places them in pinned memory for direct access by GPU kernels.

As shown in Figure~\ref{fig:overview}, this lookup path implements the \textit{Retrieve} operation. \circled{\textsf{1}} Given the indices of critical partitions, the buffer manager first queries the Meta-partition table to check their residency state and derive each partition's page count from its token count. \circled{\textsf{2}} It then consults the GPU page table to locate GPU-resident pages and perform cache replacement to make room for missed partitions, as detailed in \S~\ref{subsec:locality}. \circled{\textsf{3}} For missed partitions, the kernel reads each partition's starting offset from the Partition Offset table and fetches the corresponding contiguous slice of the CPU page table to obtain the physical CPU page IDs. \circled{\textsf{4}} It then issues PCIe transfers to bring the these pages into GPU memory.


\subsection{Dynamic, Locality-aware KV Management}
\label{subsec:locality}
\sysname's GPU buffering relies on two complementary techniques: \textit{buffer elasticity}, which dynamically resizes each request's GPU buffer based on the system load of online serving, and \textit{per-head cache replacement}, which applies a GPU-friendly bucketed LRU policy. The scheduler drives the former at each decoding step, while the latter runs within the step for every KV head.

\minisec{Buffer Elasticity}
Such elasticity aims to maximize overall serving throughput by balancing per-request throughput and system-wide concurrency. Over-allocating GPU buffer space to a request improves cache hits but reduces the number of concurrently serving requests; under-allocation has the opposite effect, increasing concurrency at the cost of lower per-request throughput. The scheduler must determine the buffer size that balances these competing effects.

\sysname makes this tradeoff explicit by splitting a request's GPU allocation into \textit{mandatory pages} and \textit{buffering pages}. Mandatory pages hold the critical partitions required for the current decoding step, whereas buffering pages retain previously retrieved partitions to exploit temporal locality.
We observe a stable sweet spot in buffer size: increasing the buffer up to this point materially improves cache hit ratios, while additional pages yield diminishing returns. 
This observation is supported by prior works~\cite{chenRetroInferVectorStorageApproach2025} and an empirical study across sparse algorithms and tasks (\S~\ref{subsec:ablation}).
\sysname therefore empirically sets the minimum buffer allocation to 5$\times$ of the mandatory pages. 

This tradeoff determines the scheduler's GPU memory allocation policy. A new request is admitted into the batch only if the system can guarantee both its mandatory pages and the minimum buffer allocation; otherwise, the request remains queued. When a new admission or increasing demand requires additional pages, the scheduler reclaims buffer pages from active requests, with each request contributing in proportion to its sequence length. We reclaim pages from the tail of the GPU page table to keep the table compact, and also fits the bucketed LRU policy described next.

\minisec{Bucketed LRU Replacement}
Within each request's GPU buffer, \sysname performs cache replacement to decide which partitions to retain and which to evict. Because this logic runs for every layer and KV head, it must be lightweight to avoid the replacement overhead erasing the benefit of cache hits. \sysname therefore keeps the control logic on the GPU and uses a GPU-friendly replacement policy tailored to sparse access patterns.

Unlike traditional policies such as LRU, \sysname relaxes the ordering and approximates LRU with bounded recency timestamps, avoiding expensive global sorting and ordering. 
Specifically, each GPU-resident page carries a timestamp drawn from a small fixed range, such as 64 values. 
This design matches sparse attention's bulk-access pattern: each step selects a batch of critical partitions, and the system needs not maintain a strict recency order among pages within the same batch. 
Therefore, using decoding step as the time unit, 
a 64-value timestamp space can distinguish partitions accessed over the last 64 decoding steps, which empirically captures the locality window for most sparse algorithms and workloads.
We name it \textit{bucketed LRU}, where each timestamp value defines a bucket of pages with similar recency.


The replacement proceeds in four steps. 
\circled{\textsf{1}} \textbf{Hit/\linebreak Miss Classification}: the buffer manager queries the Meta-partition Table to identify resident partitions (hits), non-resident partitions (misses), and the total number of GPU pages required by the misses, which we call the \textit{page demand}. 
\circled{\textsf{2}} \textbf{Timestamp Update}: each GPU-resident page stores a timestamp in the GPU page table with range $[0, n-1]$, where larger values denote more recent access. The system promotes pages of hit partitions to $\min(\mathit{decoding\_step}, n-1)$. Once the decoding step exceeds $n-1$, promoted pages remain at $n-1$ and all other pages are demoted by one. 
\circled{\textsf{3}} \textbf{Eviction}: while updating timestamps, the system scans the GPU page table once to build a shared-memory histogram of page counts per timestamp bucket. It then searches the buckets from oldest to newest to find the smallest threshold $x$ whose cumulative count satisfies the page demand, and evicts pages with timestamps at or below $x$. Eviction clears the corresponding GPU page table entries and marks the affected partitions as non-resident in the Meta-partition Table. 
\circled{\textsf{4}} \textbf{Admission}: the system marks the missed partitions resident and remaps the freed GPU pages to them by updating partition IDs and timestamps in the GPU page table. This step changes only metadata; the actual CPU-to-GPU KV copies are issued later after locating the source CPU pages.

In summary, bucketed LRU avoids global ordering, relying instead on simple scans and histogram updates that parallelize well on the GPU. During the step of admission, the system also records the target GPU page IDs for both hits and misses; these IDs are later used both to copy missed KV pages and to run attention on all selected partitions.


\subsection{Hierarchical Metadata}
\label{subsec:metadata}
\sysname's fine-grained, per-head KV management improves retrieval efficiency, but it also amplifies metadata overhead. Existing serving systems such as vLLM typically pre-allocate metadata arrays for a worst-case configuration. Applied to \sysname, this strategy would make the GPU page table alone scale as $\mathit{max\_batch\_size} \times \frac{\mathit{max\_context\_length}}{\mathit{page\_size}} \times \mathit{num\_layers} \times \mathit{num\_heads}$. Relative to conventional serving systems, the footprint grows along three axes: per-head management adds a $\mathit{num\_heads}\times\mathit{num\_layers}$ factor, extending KV capacity into CPU memory raises the maximum supported batch size, and longer context windows increase the number of logical pages per request. Other structures, such as the Meta-partition Table and Partition Offset Table, add further overhead. In aggregate, these metadata structures can consume tens or even hundreds of gigabytes of GPU memory, reducing space for cached KVs and limiting scalability.

\begin{figure}[t]
    \centering
    \includegraphics[width=\linewidth]{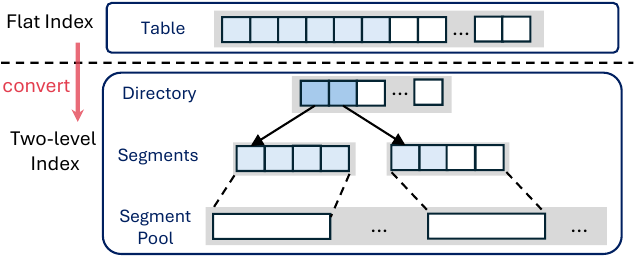}
    \caption{Two-level page table design.}
    \label{fig:metadata_hierarchy}
\end{figure}

Tier-split mapping removes part of this pressure, but it does not eliminate worst-case overprovisioning. \sysname therefore follows a principle that \textit{metadata should scale with the active physical working set rather than the worst-case logical address space}. Inspired by multi-level paging in operating systems, \sysname organizes all four metadata tables with two-level indexing. As shown in Figure~\ref{fig:metadata_hierarchy}, using the GPU page table as an example, the system keeps a small, statically allocated top-level directory on the GPU whose entries point to second-level segments. These segments are allocated \emph{on demand}, only when physical pages are assigned to a request, so metadata grows with the number of mapped physical pages rather than the worst-case logical capacity. To keep allocation inexpensive, \sysname pre-allocates a GPU-resident segment pool shared across requests and heads and manages it with a lightweight bitmap allocator. The pool is bounded by GPU memory capacity rather than the logical address space. Combined with tier-split placement, this organization reduces metadata overhead by more than an order of magnitude, as we show in \S~\ref{subsec:ablation}.

\subsection{Implementation}
\label{sec:imple}

We implement the core execution engine of \sysname in 7k lines of C++ and CUDA. To reuse mature functionality from existing serving systems, we integrate \sysname into vLLM~\cite{kwon2023efficient} with 3k lines of Python, modifying its scheduler and memory allocator to orchestrate the sparse attention pipeline. To sustain high performance, we also implement a set of optimized CUDA kernels for the key operations in the pipeline.

To efficiently transfer non-contiguous, head-wise KV pages across the GPU--CPU memory hierarchy, \sysname implements a warp-based copy kernel. Each warp independently copies one page using vectorized memory transactions, allowing the kernel as a whole to saturate PCIe bandwidth. We further optimize it with a persistent-kernel design that distributes pages across all warps regardless of their owning head or request, improving load balance.

\section{Evaluation}
\label{sec:eval}

We evaluate \sysname against state-of-the-art LLM serving engines~\cite{kwonEfficientMemoryManagement2023,yang2025lserve} and the original implementations of three representative sparse attention algorithms~\cite{sunShadowKVKVCache2025a,chenRetroInferVectorStorageApproach2025,gao2025seerattention-r}. 
Key findings include:
\begin{itemize}
    \item \sysname achieves $1.66$--$5.66\times$ higher end-to-end throughput than vLLM across models, GPUs, and workloads, with higher gains at higher request rates.

    \item \sysname's performance advantage stems from sustaining larger decoding batch sizes through reduced per-request HBM demand, while keeping prefill latency close to that of vLLM.
    
    \item \sysname reduces per-token decode latency by $8.4$--$58\%$ compared to the original implementations of the representative sparse attention algorithms, translating up to $2.39\times$ throughput improvement and thus demonstrating the effectiveness of \sysname's system-level optimizations.
    
    \item The hierachical metadata organization of \sysname reduces metadata HBM consumption by $49$--$78\times$, and locality-aware buffer management improves decode throughput by $2.18\times$.
\end{itemize}

\subsection{Experimental Setup}

We evaluate \sysname across two server platforms, three model sizes, and two long-context benchmarks.

\myparagraph{Testbed}
We conduct experiments on two servers with
different GPU architectures.
The first server has four NVIDIA A100 GPUs (80\,GB HBM each), an AMD EPYC 7V12 CPU with 850\,GB DRAM, and PCIe Gen4 links at 32\,GB/s per GPU.
The second server has four NVIDIA B200 GPUs (180\,GB HBM each), an Intel Xeon Platinum 8570 CPU with 1.5\,TB DRAM, and PCIe Gen5 links at 64\,GB/s per GPU.

\myparagraph{Models}
We use Qwen3-14B~\cite{qwen3-14b}, Qwen3-32B~\cite{qwen3-32b}, and Llama-3.1-70B~\cite{llama3.1-70B} to cover different model sizes and architectures.
Unless otherwise specified,
experiments on 1, 2, and 4 GPUs (TP=1/2/4) use Qwen3-14B, Qwen3-32B, and Llama-3.1-70B, respectively.

\myparagraph{Workloads}
We draw requests from two state-of-the-art long-context benchmarks.
LongBench-v2~\cite{longbench-v2} features long inputs and short outputs 
while LongGenBench~\cite{liu2024longgenbench} has shorter inputs but longer outputs driven by reasoning-intensive tasks, 
as shown in Table~\ref{tab:workload}.
Together they cover various tasks such as question answering and text summarization, spanning a wide range of input/output length ratios.
For online serving, we generate request arrivals with a Poisson process at varying rates, following prior work~\cite{agrawal2024taming,kwonEfficientMemoryManagement2023}.

\begin{table}[t]
    \footnotesize
    \centering
    \caption{\textbf{Workload characteristics.} LongBench-v2 has long inputs and short outputs, whereas LongGenBench exhibits shorter inputs but longer outputs. Lengths are measured in tokens.}
    \newcolumntype{C}{>{\centering\arraybackslash}X}
    \begin{tabularx}{\columnwidth}{l | C C C | C C C}
        \toprule
         & \multicolumn{3}{c|}{\textbf{Input Length}} & \multicolumn{3}{c}{\textbf{Output Length}} \\
         & \textbf{Min} & \textbf{Max} & \textbf{Avg} & \textbf{Min} & \textbf{Max} & \textbf{Avg} \\
        \midrule
        \textbf{LongBench-v2}~\cite{longbench-v2} & 32K  & 120K & 55K & 500 & 15K & 5K  \\
        \textbf{LongGenBench}~\cite{liu2024longgenbench} & 16K  & 19K  & 18K & 7K   & 32K & 12K \\
        \bottomrule
    \end{tabularx}
    \label{tab:workload}
\end{table}

\myparagraph{Baselines}
\sysname is built on vLLM~\cite{kwonEfficientMemoryManagement2023} to reuse its optimized GPU kernels and model execution infrastructure.
To demonstrate generality, we integrate three representative sparse attention methods---ShadowKV~\cite{sunShadowKVKVCache2025a}, RetroInfer~\cite{chenRetroInferVectorStorageApproach2025}, and SeerAttention-R~\cite{gao2025seerattention-r}---yielding \sysname-ShadowKV, \sysname-RetroInfer, and \sysname-SeerAttention, respectively.
Configurations of \sysname-powered systems are listed in Table~\ref{tab:sys_config}, following the default settings in their original implementations.

\begin{table}[t]
    \footnotesize
    \centering
    \caption{\textbf{Configurations of \sysname-powered systems, aligned with the defaults used in their original implementations.} ShadowKV and RetroInfer use a fixed ratio of context length as retrieval budget, while SeerAttention-R retrieves a fixed number of tokens regardless of context length. Partition granularity and physical page size are measured in tokens.}
    \newcolumntype{C}{>{\centering\arraybackslash}X}
    \setlength{\tabcolsep}{4pt}
    \begin{tabularx}{\columnwidth}{l | C | C | C}
        \toprule
         & \textbf{Retrieval Budget} & \textbf{Partition Granularity} & \textbf{Physical Page Size} \\
        \midrule
        \textbf{ShadowKV}~\cite{sunShadowKVKVCache2025a}      & $1.56\%$ & 8 & 8 \\
        \textbf{RetroInfer}~\cite{chenRetroInferVectorStorageApproach2025}    & $1.8\%$ & variable & 8 \\
        \textbf{SeerAttention-R}~\cite{gao2025seerattention-r} & 2K & 32 & 32 \\
        \bottomrule
    \end{tabularx}
    \label{tab:sys_config}
\end{table}

We compare against three baselines: (1)~vLLM with full attention; (2)~vLLM-Offload, which extends vLLM with KV-cache offloading to CPU memory under HBM pressure;
\linebreak (3)~LServe~\cite{yang2025lserve}, a GPU-only serving system that exploits attention sparsity. 
LServe is excluded from online serving experiments because its scheduler does not support online request arrivals. The original implementations of the sparse algorithms also do not support online serving due to the lack of a request scheduler and the capability of efficiently handling variable-length requests.
Therefore, we only compare with the original implementations of sparse algorithms in \S~\ref{subsec:pd} to understand the performance gain of \sysname's system optimizations. 
By default, we use Qwen3-14B on an A100 for offline performance analysis and microbenchmarks as LServe and some original sparse attention implementations only support single-GPU execution and does not have Blackwell architecture support.

\subsection{End-to-end Performance for Online Serving}
\label{subsec:e2e}

We evaluate online serving using AIPerf~\cite{ai_dynamo_aiperf_github} to quantify the end-to-end efficiency of \sysname and baselines.
\sysname-SeerAttention is evaluated only on Qwen3-14B because SeerAttention-R is training-based and currently unavailable for larger models.

\begin{figure*}[t]
    \centering
    \setlength{\abovecaptionskip}{2pt}
    \includegraphics[width=\textwidth]{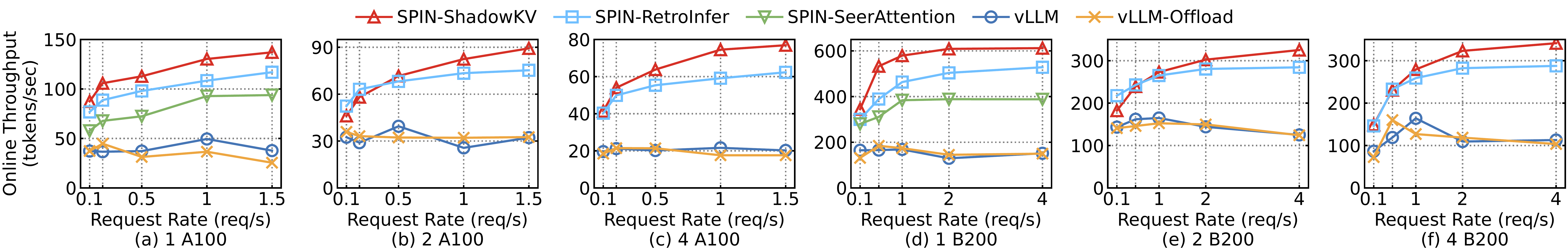}
    \caption{\textbf{End-to-end Throughput vs. Request Rate on LongBench-v2.} }
    \Description{Online serving throughput on LongBench-v2 as request rate increases, comparing \sysname-powered systems with vLLM and vLLM-Offload across different GPU counts.}
    \label{fig:online_thru_lbv2}
\end{figure*}

\begin{figure*}[t]
    \centering
    \setlength{\abovecaptionskip}{2pt}
    \includegraphics[width=\textwidth]{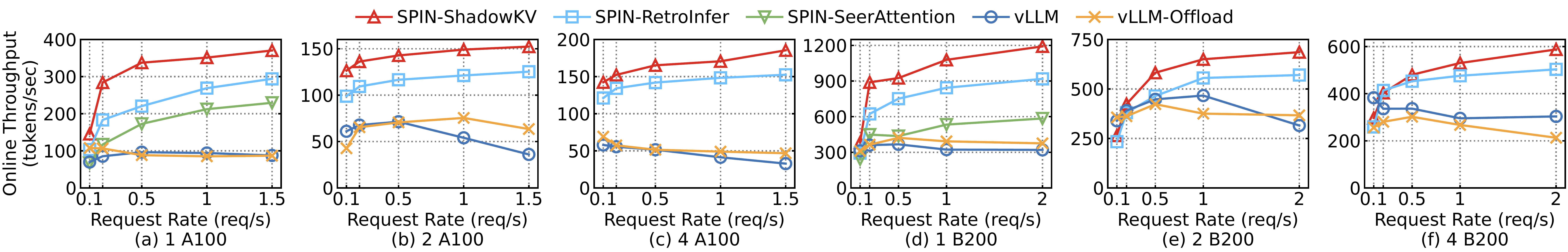}
    \caption{\textbf{End-to-end Throughput vs. Request Rate on LongGenBench.}}
    \Description{Online serving throughput on LongGenBench as request rate increases, comparing \sysname-powered systems with vLLM and vLLM-Offload across different GPU counts.}
    \label{fig:online_thru_lgb}
\end{figure*}


\begin{figure}[t]
    \centering
    \setlength{\abovecaptionskip}{2pt}
    \includegraphics[width=0.95\linewidth]{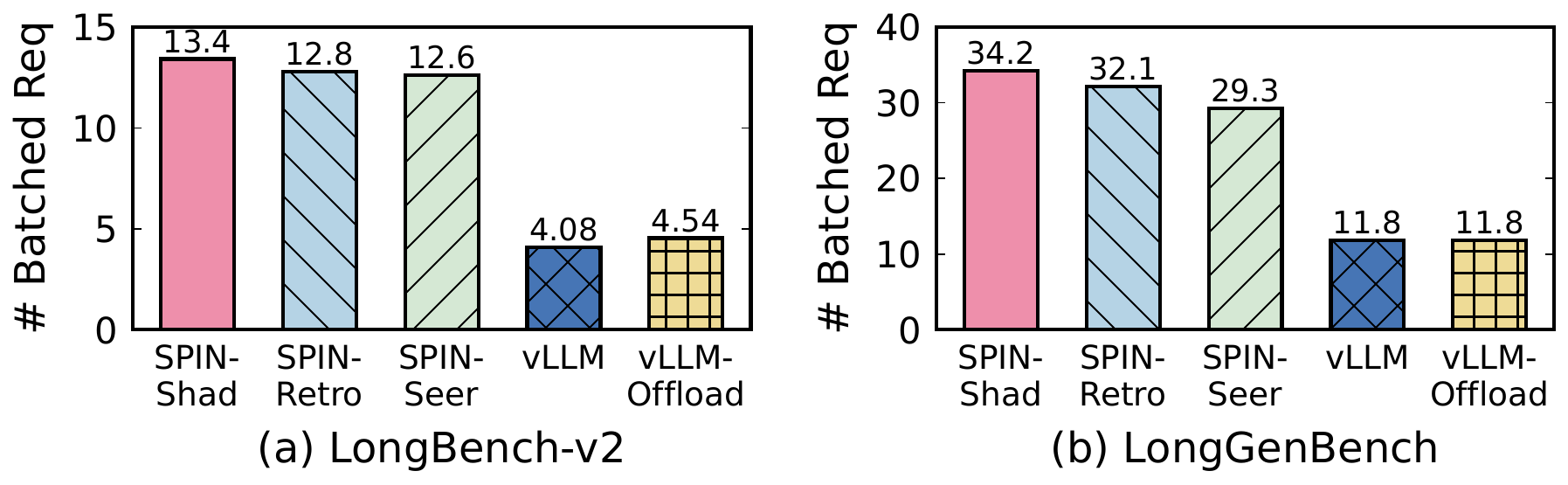}
    \caption{\textbf{Average Batch Size.} Average number of batched requests when serving Qwen3-14B on an A100 for the LongBench-v2 and LongGenBench workloads (both at 1.5 req/s).}
    \label{fig:batch_size}
\end{figure}

\begin{figure}[t]
    \centering
    \setlength{\abovecaptionskip}{2pt}
    \includegraphics[width=\linewidth]{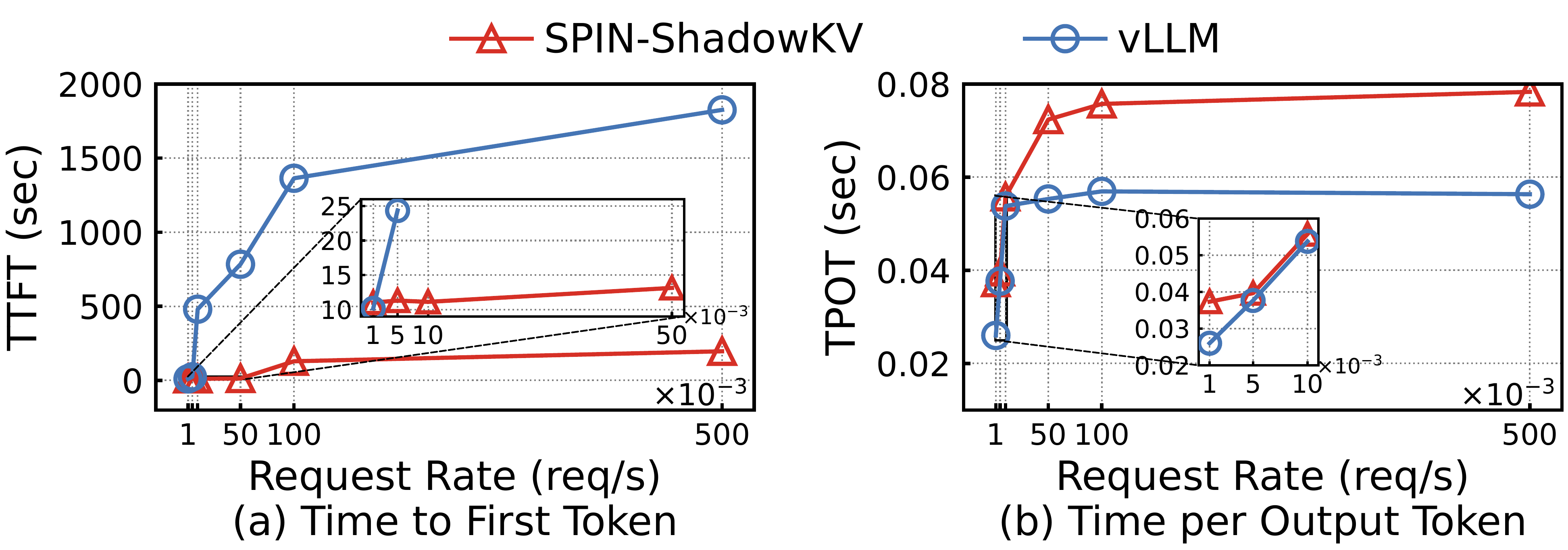}
    \caption{\textbf{Average Latency.} Average TTFT and TPOT when serving Qwen3-14B on an A100 for the LongBench-v2 workloads under varying request rates.}
    \label{fig:online_ttft_tpot}
\end{figure}

\myparagraph{Throughput}
Figures~\ref{fig:online_thru_lbv2} and~\ref{fig:online_thru_lgb} report end-to-end output-token throughput, defined as the total number of output tokens divided by the system serving time, under varying request rates.
Across all workloads and GPU platforms, \sysname-powered systems consistently translate attention sparsity into substantial throughput gains.
On LongBench-v2 (Figure~\ref{fig:online_thru_lbv2}), \sysname-powered systems continue to scale with request rate, while vLLM and vLLM-Offload quickly plateau and even decline under high load due to memory thrashing.
By reducing per-request HBM demand through effective GPU--CPU memory management, \sysname sustains larger running batches and thus better utilize GPU resources, achieving $2.34$--$3.80\times$ higher throughput than vLLM on A100 at 1.5\,req/s and $2.27$--$4.03\times$ on B200 at 4\,req/s.
Figure~\ref{fig:batch_size} confirms that \sysname-powered systems support $2.5$--$3.3\times$ larger average batch sizes than vLLM and vLLM-Offload during online serving.
Among the three sparse variants, \sysname-SeerAttention delivers relatively lower throughput because its larger per-step retrieval budget required by algorithm incurs higher PCIe transfer overhead.

Figure~\ref{fig:online_thru_lgb} for LongGenBench shows a similar trend, with \sysname-powered systems achieving $2.62$--$5.66\times$ higher throughput than vLLM on A100 at 1.5 req/s, and $1.66$--$3.72\times$ higher throughput on B200 at 4 req/s.
These results demonstrate that \sysname provides robust end-to-end throughput gains across workloads with different input/output length profiles.

\myparagraph{Latency}
Figure~\ref{fig:online_ttft_tpot} reports average time-to-first-token (TTFT) and time-per-output-token (TPOT) for Qwen3-14B on A100 under varying request rates; for clarity, we compare only \sysname-ShadowKV with vLLM.
At low load ($\leq 0.001$\,req/s), both systems exhibit comparable TTFT because their prefill times are similar and request queueing is rare.
As request rate grows, \sysname-ShadowKV achieves $9\times$ lower TTFT than vLLM at 0.5\,req/s, because \sysname sustains larger batch sizes and higher throughput and thereby reduces queueing delay for newly arrived requests.
\sysname incurs higher TPOT than vLLM due to larger per-step decode batches and GPU--CPU data transfer for retrieval.
This gap, however, remains stable across request rates, as it is governed by the minimum buffer size per request.
Other \sysname variants exhibit the same trend: \sysname-SeerAttention and \sysname-RetroInfer add $107\%$ and $14.7\%$ TPOT over \sysname-ShadowKV, respectively, 
due to algorithmic differences in retrieval budget.

\subsection{Offline Performance}
\label{subsec:pd}

\begin{figure}[t]
    \centering
    \setlength{\abovecaptionskip}{2pt}
    \includegraphics[width=0.83\linewidth]{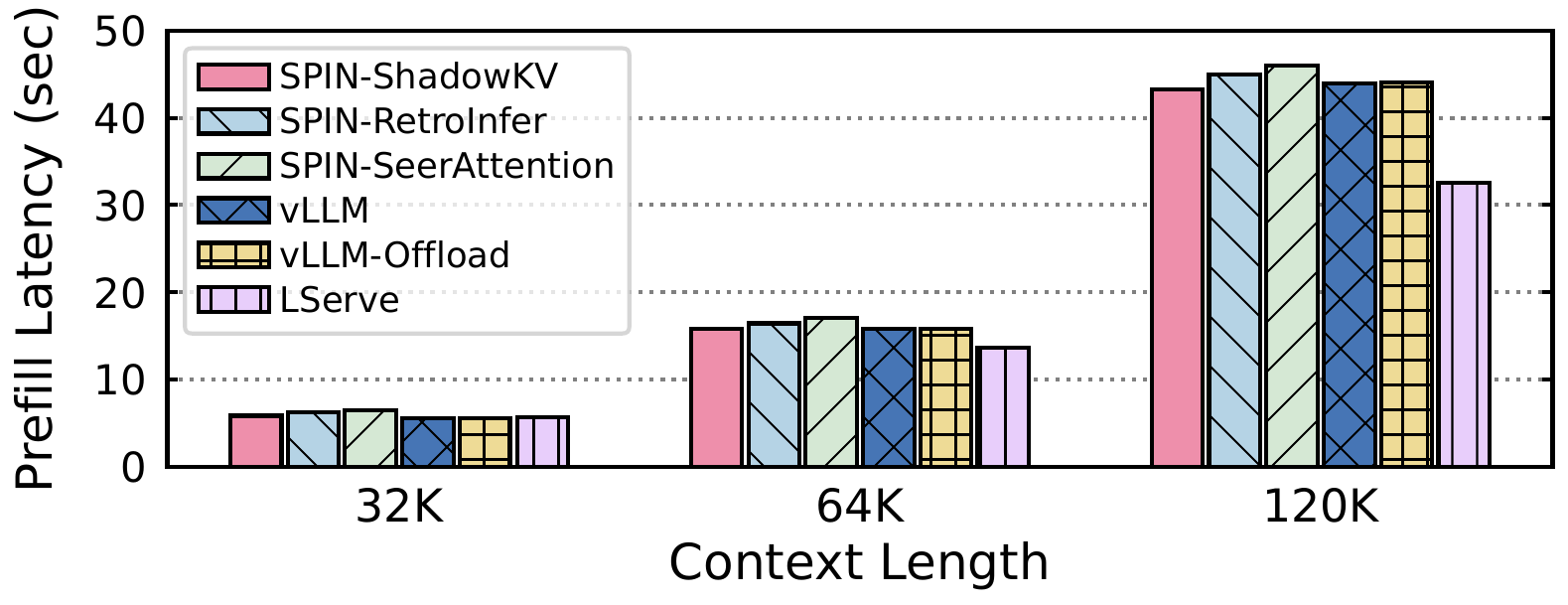}
    \caption{\textbf{Prefill Latency vs. Context Length on an A100.} 
    }
    \Description{Prefill latency on A100 and B200 GPUs across different context lengths for batch size one, comparing \sysname-powered systems with baseline systems.}
    \label{fig:offline_prefill_latency}
\end{figure}

\begin{figure*}[t]
    \centering
    \setlength{\abovecaptionskip}{2pt}
    \includegraphics[width=\textwidth]{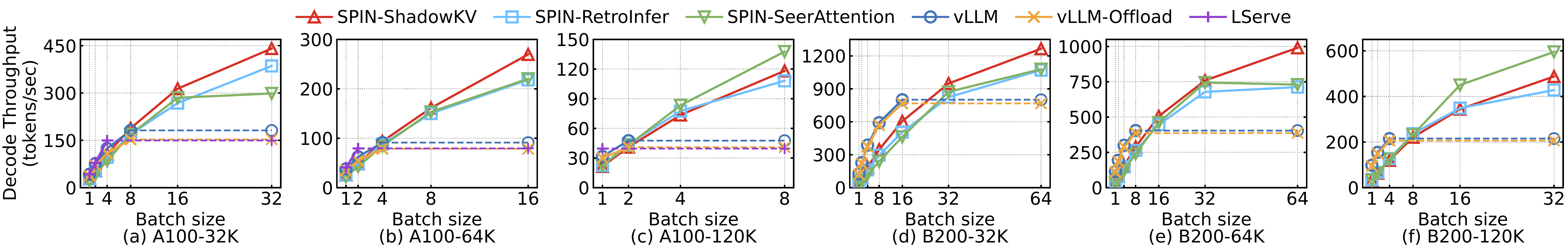}
    \caption{\textbf{Offline Decode Throughput vs. Batch Size on different Context Lengths and GPUs.} 
    Dashed horizontal lines extend each baseline’s throughput at its maximum supported batch size for ease of comparison.
    }
    \Description{Decode throughput of offline serving on A100 and B200 GPUs across multiple context lengths and batch sizes, comparing \sysname-powered sparse-attention systems with vLLM and vLLM-Offload.\bt{Remove the results on 16K context length}}
    \label{fig:offline_decode_throughput}
\end{figure*}

We evaluate offline inference with Qwen3-14B, measuring prefill and decode separately across context lengths to enable stage-level analysis.
This controlled setting also permits direct comparison with LServe and the original sparse implementations, which do not support online scheduling.

\myparagraph{Prefill} 
Figure~\ref{fig:offline_prefill_latency} shows prefill latency across context lengths.
At 16K, 32K, and 64K contexts, \sysname-powered systems add at most 0.59, 0.92, and 1.32\,s over vLLM, respectively, because \sysname's indexing and KV offloading execute largely asynchronously with GPU prefill computation.
In some configurations \sysname even matches or undercuts vLLM's prefill latency, owing to its GPU-side memory allocator that avoids the host-side allocation overhead at long contexts.
LServe achieves $2.5$--$29\%$ lower prefill latency than \sysname by applying sparse prefill to reduce attention computation; this algorithmic optimization is orthogonal to \sysname and can be integrated into our system for further improvement.

\myparagraph{Decode}
Figure~\ref{fig:offline_decode_throughput} evaluates offline decode throughput as batch sizes increases under different context lengths.
\sysname-powered systems support $4$--$8\times$ larger maximum batch sizes than baselines, translating into $1.34$--$3.49\times$ higher throughput.
LServe improves over vLLM by exploiting decode-stage sparsity on GPU, but its GPU-only design caps its maximum batch size.
Among \sysname variants, \sysname-SeerAttention delivers the highest decode throughput at 120K context because its fixed retrieval budget is independent of context length, whereas \sysname-ShadowKV and \sysname-RetroInfer require retrieving proportionally more tokens as context grows.
On B200 shown in Figure~\ref{fig:offline_decode_throughput}(d--f), \sysname's 
decode throughput at small batch sizes falls below vLLM, because B200's higher compute and HBM bandwidth relatively alleviates the full-attention bottleneck.
Nevertheless, \sysname maintains its advantage at larger batch sizes by leveraging hierarchical GPU--CPU memory to sustain higher concurrency. 

\begin{figure}[t]
    \centering
    \setlength{\abovecaptionskip}{2pt}
    \includegraphics[width=\linewidth]{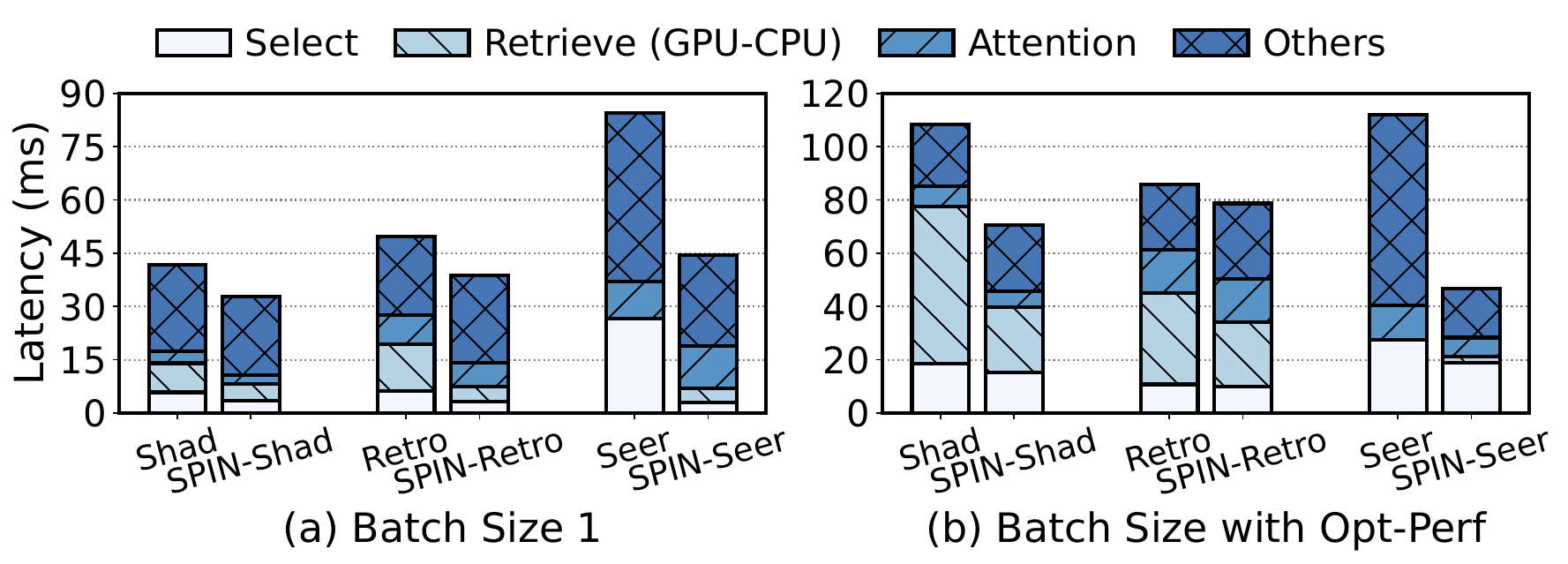}
    \caption{\textbf{Per-token Decode Latency Breakdown of \sysname and Original Sparse Implementations.}
    \sysname consistently reduces per-token decode latency across all three sparse attention algorithms, 
    where gains primarily come from lower PCIe retrieval overhead for ShadowKV and RetroInfer and from more efficient GPU kernels for SeerAttention-R.
    }
    \label{fig:eval:breakdown_spin_vs_ori}
\end{figure}

\myparagraph{Comparison with Original Sparse Implementations}
We compare \sysname variants against the original implementations of their respective sparse attention algorithms to analyze the gains from \sysname's system-level optimizations.
As shown in Figure~\ref{fig:eval:breakdown_spin_vs_ori}, \sysname-ShadowKV, \sysname-RetroInfer, and \sysname-SeerAttention reduce per-token decode latency by $21\%$, $22\%$, and $47\%$ over their original counterparts at batch size 1, and by $35\%$, $8.4\%$, and $58\%$ at their respective peak batch sizes.
The latency breakdown reveals that the dominant source of improvement differs across algorithms.
For ShadowKV and RetroInfer, \sysname's locality-aware buffer manager significantly reduces PCIe retrieval time by exploiting the temporal locality of token accesses.
For SeerAttention-R, whose original implementation runs entirely on GPU and thus incurs no retrieval overhead, the gains instead come from \sysname's optimized GPU kernels that replace the under-optimized operators in the original codebase. Overall, the latency improvement translates up to $2.39\times$ throughput gains compared to the original implementations.

\subsection{Ablation Studies}
\label{subsec:ablation}

We quantify the impact of individual design decisions of \sysname including the effect of GPU buffer management, multi-level indexing, and cache hit ratios under different buffer sizes. 
Unless otherwise noted, ablations use \sysname-ShadowKV with Qwen3-14B at 32K context on A100.

\begin{figure}[tbp]
  \centering
  \begin{minipage}[t]{0.5\linewidth}
    \centering
    \setlength{\abovecaptionskip}{2pt}
    \includegraphics[width=\textwidth]{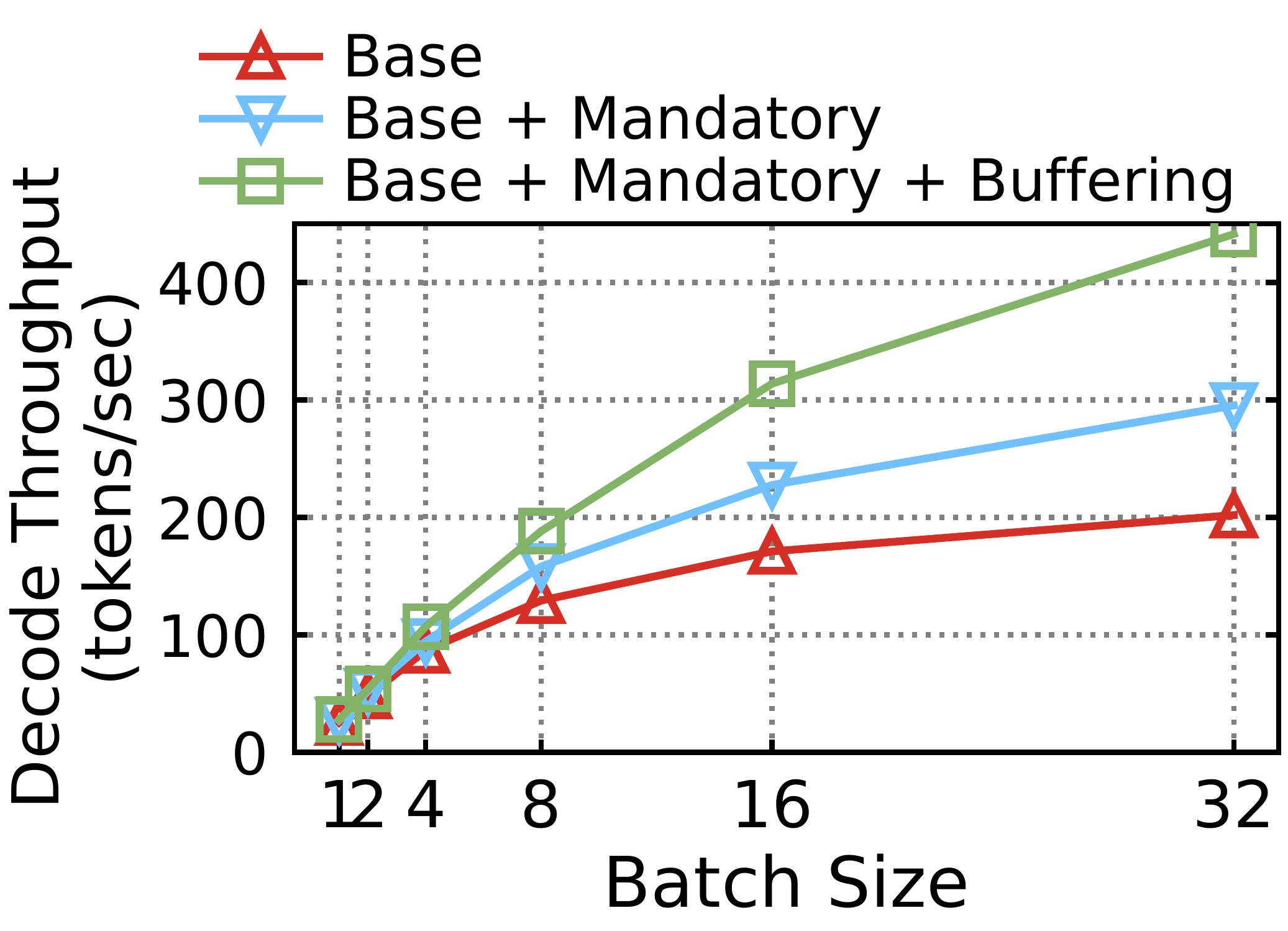}
        \caption{\textbf{Effect of Design Choices in Buffer Management.}
        ``Base'' offloads all KV to CPU without GPU caching.
        ``+\,Mandatory'' reuses visited pages only from the last decode step. 
        ``+\,Mandatory\,+\,Buffering'' fully exploits cross-step temporal locality via bucketed LRU.
        }
        \label{fig:ablation_dec_thru}
        
  \end{minipage}
  \hspace{0em} 
  \begin{minipage}[t]{0.48\linewidth}
    \centering
    \setlength{\abovecaptionskip}{2pt}
    \includegraphics[width=\textwidth]{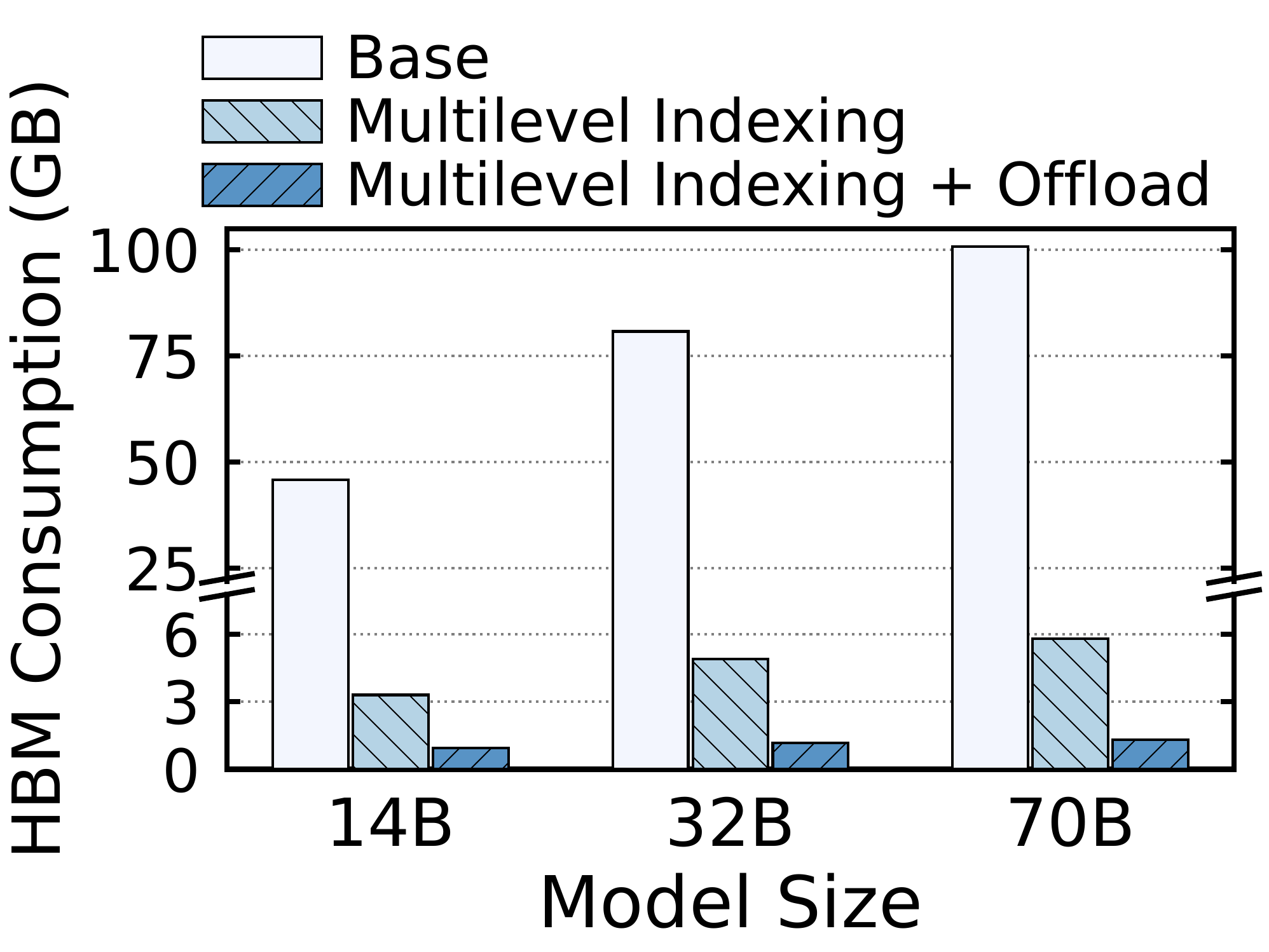}
            \caption{\textbf{HBM Consumption of Metadata for Buffer Management.}
            ``Base'' maintains the flat page tables and all related metadata (e.g., partition table) in GPU memory. 
            Multilevel indexing and CPU offloading reduce orders of magnitude HBM consumption.}
       \label{fig:ablation_metadata_hbm}
  \end{minipage}
\end{figure}

\myparagraph{Effect of GPU Buffer Management}
Figure~\ref{fig:ablation_dec_thru} compares three incremental configurations to quantify the benefit of \sysname's GPU buffer management.
Without GPU caching (``Base''), decode throughput plateaus at low batch sizes because every retrieval must traverse PCIe, saturating link bandwidth.
Retaining pages from the last decode step on GPU (``+\,Mandatory'') reduces transfer volume and improves throughput, but captures only single-step reuse and misses locality that spans multiple steps.
Adding bucketed LRU fully exploits cross-step temporal locality, raising decode throughput by $49\%$ by keeping frequently accessed pages resident and eliminating redundant transfers.

\begin{figure}[t]
    \centering
    \setlength{\abovecaptionskip}{2pt}
    \includegraphics[width=0.95\linewidth]{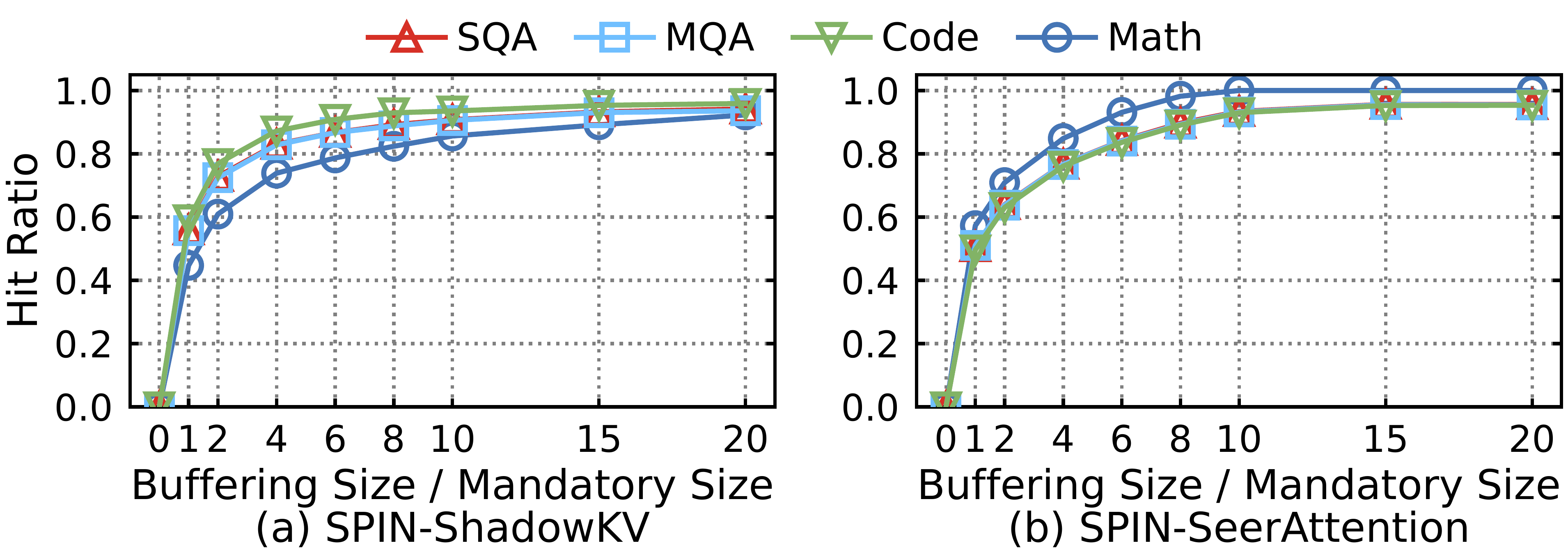}
    \caption{\textbf{Cache Hit Ratio vs. Cache Size.} Increasing cache size significantly improves hit ratio; however, further increases beyond a certain point yield only marginal improvement.}
    \label{fig:cache_hit_ratio}
\end{figure}

\myparagraph{Impact of Cache Sizes}
We study how cache size affects hit ratio across LongBench-v2 and LongGenBench tasks and sparse attention algorithms.
As shown in Figure~\ref{fig:cache_hit_ratio}, enlarging the cache significantly improves hit ratio, confirming strong temporal locality in the critical tokens accessed during decoding.
This gain saturates once the Buffering-to-Mandatory ratio reaches $4\times$.
Beyond this point, additional cache capacity yields only marginal hit-ratio improvement while consuming more GPU memory, which could in turn reduce the maximum executable batch size and therefore limit throughput scalability.
We observe the same trend across workloads, sparse attention algorithms, and models.
These results guide the scheduler's cache-size setting: a moderate cache size strikes the best balance between hit ratio and GPU memory consumption, allowing the system to admit more requests without incurring excessive retrieval overhead.

\myparagraph{Multi-level Indexing}
We next quantify the HBM savings provided by \sysname's multi-level indexing and tier-plit design across different models.
As shown in Figure~\ref{fig:ablation_metadata_hbm}, a naive design that maintains a flat page table and related metadata for the full working set (128K context and batch size 32) consumes up to 100\,GB of HBM for Llama-3.1-70B.
Replacing this flat layout with a two-level index that partitions the page table and related metadata into segments reduces metadata HBM consumption by $13.9$--$17.4\times$.
Offloading selected tables, including CPU partition offset tables and page tables, to CPU memory further reduces GPU memory usage by $3.5$--$4.5\times$, freeing HBM for KV cache storage and thereby enabling larger batch sizes and higher throughput.

\section{Related Work}
\label{sec:related}

\myparagraph{LLM Serving Optimizations}
A rich body of work has improved LLM serving throughput through better batching~\cite{yu2022orca,agrawal2023sarathi}, memory management~\cite{kwon2023efficient, zhangJengaEffectiveMemory2025a, prabhuVAttentionDynamicMemory2025}, scheduling~\cite{zheng2024sglang, meiHelixServingLarge2025,zhong2024distserve, patelSplitwiseEfficientGenerative2024, sunLlumnixDynamicScheduling2024} and parallelism~\cite{wu2024loongserve, liu2023ringattentionblockwisetransformers, jacobs2023deepspeedulyssesoptimizationsenabling}. To efficiently handle long-context workloads, several lines of optimization have emerged.
Speculative decoding~\cite{cai2024medusasimplellminference, li2025eaglespeculativesamplingrequires, li2025eagle3scalinginferenceacceleration} reduces latency by predicting tokens via a lightweight draft model with verification. Offloading-based systems~\cite{qin2025mooncake, liu2025lmcacheefficientkvcache, sheng2023flexgen, lin2024infinitellmefficientllmservice} move KV cache or model weights across a multi-tier memory hierarchy to fit larger contexts.
These studies do not address sparse attention, thereby omitting the fundamental abstractions and kernels necessary for efficient serving.

\myparagraph{KV Cache Pruning and Quantization}
KV pruning methods~\cite{h2o, streamingllm, li2024snapkv, wan2024d2o} permanently remove less critical tokens from the KV cache, whereas KV quantization techniques~\cite{hooper2024kvquant, liu2024kivi, zhang2024kv} compress the existing entries into lower-precision formats. Both approaches assume inherent redundancy within the KV cache and are thus not \sysname's target algorithms.

\myparagraph{Sparse Attention}
Rather than permanently evicting tokens, sparse attention methods~\cite{sunShadowKVKVCache2025a, chenArkValeEfficientGenerative2024, chenRetroInferVectorStorageApproach2025, liu2024retrievalattention, magicpig, xiao2024duoattention, gao2025seerattention-r, yuan2025native, liu2025deepseek} retain the complete KV cache in memory but dynamically select a critical subset of tokens for attention computation during each decoding step. These methods observe token importance varies as generation progresses and thus exploit it for better accuracy. Other works~\cite{jiang2024minference,yang2025lserve, xuxattention} exploit sparsity during the prefill phase, complementary to decode-phase sparsity.
\sysname can support both lines of work with minimal adaptation.
While LServe~\cite{yang2025lserve} also supports both prefill and decode sparsity, it is limited to the specific sparse attention algorithms. In contrast, \sysname is designed as a general serving engine that executes a wide range of sparse attention algorithms with different granularities at maximal efficiency.

\myparagraph{Model Quantization and Sparsity}
Weight sparsity~\cite{lin2024awq, zhao2024atom, alizadeh2024llm} compresses model parameters to reduce model size and FFN computation. Other systems~\cite{song2024powerinfer, xue2024powerinfer} exploit activation sparsity to offload cold FFN weights to CPU memory.
These are also promising lines of work to mitigate HBM contention and IO bottlenecks, which are orthogonal to \sysname's design.

\section{Conclusion}

This paper introduces \sysname, a sparse attention-native framework for efficient inference in long-context LLM serving. Through a unified pipeline abstraction, \sysname allows diverse sparse attention algorithms to be integrated with minimal developer effort --- requiring only the algorithm-defined Index and Select logic while the rest is automatically handled by the system. Built on top of this abstraction, \sysname's locality-aware buffer management and scalable metadata organization translate algorithmic sparsity into system-level throughput gains. Our evaluation demonstrates that \sysname achieves $1.66$ to $5.66\times$ higher throughput and $7$ to $9\times$ lower TTFT than vLLM across a variety of algorithms, models, workloads and GPU generations.

\bibliographystyle{ACM-Reference-Format}
\bibliography{main}

@inproceedings{kwonEfficientMemoryManagement2023,
  title      = {Efficient {{Memory Management}} for {{Large Language Model Serving}} with {{PagedAttention}}},
  booktitle  = {29th {{Symposium}} on {{Operating Systems Principles}}},
  author     = {Kwon, Woosuk and Li, Zhuohan and Zhuang, Siyuan and Sheng, Ying and Zheng, Lianmin and Yu, Cody Hao and Gonzalez, Joseph and Zhang, Hao and Stoica, Ion},
  year       = {2023},
  date       = {2023-10-23},
  pages      = {611--626},
  publisher  = {ACM},
  location   = {Koblenz Germany},
  doi        = {10.1145/3600006.3613165},
  url        = {https://dl.acm.org/doi/10.1145/3600006.3613165},
  eventtitle = {{{SOSP}} '23: 29th {{Symposium}} on {{Operating Systems Principles}}},
  isbn       = {979-8-4007-0229-7},
  langid     = {english}
}

@inproceedings{zhangJengaEffectiveMemory2025a,
  title = {Jenga: {{Effective Memory Management}} for {{Serving LLM}} with {{Heterogeneity}}},
  shorttitle = {Jenga},
  booktitle = {Proceedings of the {{ACM SIGOPS}} 31st {{Symposium}} on {{Operating Systems Principles}}},
  author = {Zhang, Chen and Du, Kuntai and Liu, Shu and Kwon, Woosuk and Mo, Xiangxi and Wang, Yufeng and Liu, Xiaoxuan and You, Kaichao and Li, Zhuohan and Long, Mingsheng and Zhai, Jidong and Gonzalez, Joseph and Stoica, Ion},
  year = 2025,
  month = oct,
  series = {{{SOSP}} '25},
  pages = {446--461},
  publisher = {Association for Computing Machinery},
  address = {New York, NY, USA},
  doi = {10.1145/3731569.3764823},
  urldate = {2025-10-18},
  isbn = {979-8-4007-1870-0},
}

@inproceedings{prabhuVAttentionDynamicMemory2025,
  title = {{{vAttention}}: {{Dynamic Memory Management}} for {{Serving LLMs}} without {{PagedAttention}}},
  shorttitle = {{{vAttention}}},
  booktitle = {Proceedings of the 30th {{ACM International Conference}} on {{Architectural Support}} for {{Programming Languages}} and {{Operating Systems}}, {{Volume}} 1},
  author = {Prabhu, Ramya and Nayak, Ajay and Mohan, Jayashree and Ramjee, Ramachandran and Panwar, Ashish},
  date = {2025-03-30},
  series = {{{ASPLOS}} '25},
  pages = {1133--1150},
  publisher = {Association for Computing Machinery},
  location = {New York, NY, USA},
  doi = {10.1145/3669940.3707256},
  url = {https://dl.acm.org/doi/10.1145/3669940.3707256},
  urldate = {2025-04-08},
  isbn = {979-8-4007-0698-1},
}

@inproceedings{sunLlumnixDynamicScheduling2024,
  title = {Llumnix: {{Dynamic Scheduling}} for {{Large Language Model Serving}}},
  shorttitle = {Llumnix},
  booktitle = {18th {{USENIX Symposium}} on {{Operating Systems Design}} and {{Implementation}} ({{OSDI}} 24)},
  author = {Sun, Biao and Huang, Ziming and Zhao, Hanyu and Xiao, Wencong and Zhang, Xinyi and Li, Yong and Lin, Wei},
  date = {2024-07},
  pages = {173--191},
  publisher = {USENIX Association},
  location = {Santa Clara, CA},
  url = {https://www.usenix.org/conference/osdi24/presentation/sun-biao},
  isbn = {978-1-939133-40-3},
  langid = {english},
}

@inproceedings{meiHelixServingLarge2025,
  title = {Helix: {{Serving Large Language Models}} over {{Heterogeneous GPUs}} and {{Network}} via {{Max-Flow}}},
  shorttitle = {Helix},
  booktitle = {Proceedings of the 30th {{ACM International Conference}} on {{Architectural Support}} for {{Programming Languages}} and {{Operating Systems}}, {{Volume}} 1},
  author = {Mei, Yixuan and Zhuang, Yonghao and Miao, Xupeng and Yang, Juncheng and Jia, Zhihao and Vinayak, Rashmi},
  date = {2025-02-06},
  series = {{{ASPLOS}} '25},
  pages = {586--602},
  publisher = {Association for Computing Machinery},
  location = {New York, NY, USA},
  doi = {10.1145/3669940.3707215},
  url = {https://doi.org/10.1145/3669940.3707215},
  urldate = {2025-02-22},
  isbn = {979-8-4007-0698-1},
}

@inproceedings{patelSplitwiseEfficientGenerative2024,
  title = {Splitwise: {{Efficient Generative LLM Inference Using Phase Splitting}}},
  shorttitle = {Splitwise},
  booktitle = {2024 {{ACM}}/{{IEEE}} 51st {{Annual International Symposium}} on {{Computer Architecture}} ({{ISCA}})},
  author = {Patel, Pratyush and Choukse, Esha and Zhang, Chaojie and Shah, Aashaka and Goiri, Íñigo and Maleki, Saeed and Bianchini, Ricardo},
  date = {2024-06},
  pages = {118--132},
  doi = {10.1109/ISCA59077.2024.00019},
  url = {https://ieeexplore.ieee.org/document/10609649/},
  urldate = {2025-04-22},
  eventtitle = {2024 {{ACM}}/{{IEEE}} 51st {{Annual International Symposium}} on {{Computer Architecture}} ({{ISCA}})},
}

@misc{cai2024medusasimplellminference,
  title={Medusa: Simple LLM Inference Acceleration Framework with Multiple Decoding Heads}, 
  author={Tianle Cai and Yuhong Li and Zhengyang Geng and Hongwu Peng and Jason D. Lee and Deming Chen and Tri Dao},
  year={2024},
  eprint={2401.10774},
  archivePrefix={arXiv},
  primaryClass={cs.LG},
  url={https://arxiv.org/abs/2401.10774}, 
}

@misc{li2025eaglespeculativesamplingrequires,
  title={EAGLE: Speculative Sampling Requires Rethinking Feature Uncertainty}, 
  author={Yuhui Li and Fangyun Wei and Chao Zhang and Hongyang Zhang},
  year={2025},
  eprint={2401.15077},
  archivePrefix={arXiv},
  primaryClass={cs.LG},
  url={https://arxiv.org/abs/2401.15077}, 
}

@misc{li2025eagle3scalinginferenceacceleration,
  title={EAGLE-3: Scaling up Inference Acceleration of Large Language Models via Training-Time Test}, 
  author={Yuhui Li and Fangyun Wei and Chao Zhang and Hongyang Zhang},
  year={2025},
  eprint={2503.01840},
  archivePrefix={arXiv},
  primaryClass={cs.CL},
  url={https://arxiv.org/abs/2503.01840}, 
}

@misc{liu2025lmcacheefficientkvcache,
  title={LMCache: An Efficient KV Cache Layer for Enterprise-Scale LLM Inference}, 
  author={Yuhan Liu and Yihua Cheng and Jiayi Yao and Yuwei An and Xiaokun Chen and Shaoting Feng and Yuyang Huang and Samuel Shen and Rui Zhang and Kuntai Du and Junchen Jiang},
  year={2025},
  eprint={2510.09665},
  archivePrefix={arXiv},
  primaryClass={cs.LG},
  url={https://arxiv.org/abs/2510.09665}, 
}

@misc{lin2024infinitellmefficientllmservice,
  title={Infinite-LLM: Efficient LLM Service for Long Context with DistAttention and Distributed KVCache}, 
  author={Bin Lin and Chen Zhang and Tao Peng and Hanyu Zhao and Wencong Xiao and Minmin Sun and Anmin Liu and Zhipeng Zhang and Lanbo Li and Xiafei Qiu and Shen Li and Zhigang Ji and Tao Xie and Yong Li and Wei Lin},
  year={2024},
  eprint={2401.02669},
  archivePrefix={arXiv},
  primaryClass={cs.DC},
  url={https://arxiv.org/abs/2401.02669}, 
}

@misc{liu2023ringattentionblockwisetransformers,
  title={Ring Attention with Blockwise Transformers for Near-Infinite Context}, 
  author={Hao Liu and Matei Zaharia and Pieter Abbeel},
  year={2023},
  eprint={2310.01889},
  archivePrefix={arXiv},
  primaryClass={cs.CL},
  url={https://arxiv.org/abs/2310.01889}, 
}

@misc{jacobs2023deepspeedulyssesoptimizationsenabling,
  title={DeepSpeed Ulysses: System Optimizations for Enabling Training of Extreme Long Sequence Transformer Models}, 
  author={Sam Ade Jacobs and Masahiro Tanaka and Chengming Zhang and Minjia Zhang and Shuaiwen Leon Song and Samyam Rajbhandari and Yuxiong He},
  year={2023},
  eprint={2309.14509},
  archivePrefix={arXiv},
  primaryClass={cs.LG},
  url={https://arxiv.org/abs/2309.14509}, 
}

@misc{chenRetroInferVectorStorageApproach2025,
  title       = {{{RetroInfer}}: {{A Vector-Storage Approach}} for {{Scalable Long-Context LLM Inference}}},
  shorttitle  = {{{RetroInfer}}},
  author      = {Chen, Yaoqi and Zhang, Jinkai and Lu, Baotong and Zhang, Qianxi and Zhang, Chengruidong and Luo, Jingjia and Liu, Di and Jiang, Huiqiang and Chen, Qi and Liu, Jing and Ding, Bailu and Yan, Xiao and Jiang, Jiawei and Chen, Chen and Zhang, Mingxing and Yang, Yuqing and Yang, Fan and Yang, Mao},
  year        = {2025},
  date        = {2025-05-05},
  eprint      = {2505.02922},
  eprinttype  = {arXiv},
  eprintclass = {cs},
  doi         = {10.48550/arXiv.2505.02922},
  url         = {http://arxiv.org/abs/2505.02922},
  urldate     = {2025-05-09},
  langid      = {english},
  pubstate    = {prepublished}
}

@inproceedings{sunShadowKVKVCache2025a,
  title      = {{{ShadowKV}}: {{KV Cache}} in {{Shadows}} for {{High-Throughput Long-Context LLM Inference}}},
  shorttitle = {{{ShadowKV}}},
  author     = {Sun, Hanshi and Chang, Li-Wen and Bao, Wenlei and Zheng, Size and Zheng, Ningxin and Liu, Xin and Dong, Harry and Chi, Yuejie and Chen, Beidi},
  year       = {2025},
  date       = {2025-06-18},
  url        = {https://openreview.net/forum?id=oa7MYAO6h6},
  urldate    = {2025-08-11},
  eventtitle = {Forty-Second {{International Conference}} on {{Machine Learning}}},
  langid     = {english}
}

@inproceedings{chenArkValeEfficientGenerative2024,
  title      = {{{ArkVale}}: {{Efficient Generative LLM Inference}} with {{Recallable Key-Value Eviction}}},
  shorttitle = {{{ArkVale}}},
  author     = {Chen, Renze and Wang, Zhuofeng and Cao, Beiquan and Wu, Tong and Zheng, Size and Li, Xiuhong and Wei, Xuechao and Yan, Shengen and Li, Meng and Liang, Yun},
  year       = {2024},
  date       = {2024-11-06},
  url        = {https://openreview.net/forum?id=4oAt5L4lYe&referrer=%5Bthe%20profile%20of%20Renze%20Chen%5D(%2Fprofile%3Fid%3D~Renze_Chen1)},
  urldate    = {2025-06-12},
  eventtitle = {The {{Thirty-eighth Annual Conference}} on {{Neural Information Processing Systems}}},
  langid     = {english}
}

@inproceedings{tangQUESTQueryAwareSparsity2024,
  title      = {{{QUEST}}: {{Query-Aware Sparsity}} for {{Efficient Long-Context LLM Inference}}},
  shorttitle = {{{QUEST}}},
  booktitle  = {Forty-First {{International Conference}} on {{Machine Learning}}},
  author     = {Tang, Jiaming and Zhao, Yilong and Zhu, Kan and Xiao, Guangxuan and Kasikci, Baris and Han, Song},
  year       = 2024,
  month      = jun,
  urldate    = {2025-05-14},
  langid     = {english}
}

@article{belady1966study,
  title     = {A study of replacement algorithms for a virtual-storage computer},
  author    = {Belady, Laszlo A.},
  journal   = {IBM Systems journal},
  volume    = {5},
  number    = {2},
  pages     = {78--101},
  year      = {1966},
  publisher = {IBM}
}

@article{tokenselect,
  author  = {Wei Wu and
             Zhuoshi Pan and
             Chao Wang and
             Liyi Chen and
             Yunchu Bai and
             Kun Fu and
             Zheng Wang and
             Hui Xiong},
  title   = {{TokenSelect}: Efficient Long-Context Inference and Length Extrapolation for LLMs via Dynamic Token-Level {KV} Cache Selection},
  journal = {CoRR},
  volume  = {abs/2411.02886},
  year    = {2024},
  doi     = {10.48550/ARXIV.2411.02886}
}

@inproceedings{liu2024kivi,
  address   = {Vienna, Austria},
  author    = {Zirui Liu and
               Jiayi Yuan and
               Hongye Jin and
               Shaochen (Henry) Zhong and
               Zhaozhuo Xu and
               Vladimir Braverman and
               Beidi Chen and
               Xia Hu},
  title     = {{KIVI:} {A} Tuning-Free Asymmetric 2bit Quantization for {KV} Cache},
  booktitle = {Forty-first International Conference on Machine Learning},
  publisher = {OpenReview.net},
  year      = {2024},
  url       = {https://openreview.net/forum?id=L057s2Rq8O}
}

@inproceedings{sheng2023flexgen,
  address   = {Honolulu, Hawaii, {USA}},
  author    = {Ying Sheng and
               Lianmin Zheng and
               Binhang Yuan and
               Zhuohan Li and
               Max Ryabinin and
               Beidi Chen and
               Percy Liang and
               Christopher R{\'{e}} and
               Ion Stoica and
               Ce Zhang},
  title     = {{FlexGen}: High-Throughput Generative Inference of Large Language Models
               with a Single {GPU}},
  booktitle = {Fortieth International Conference on Machine Learning},
  series    = {Proceedings of Machine Learning Research},
  volume    = {202},
  pages     = {31094--31116},
  publisher = {{PMLR}},
  year      = {2023},
  url       = {https://proceedings.mlr.press/v202/sheng23a.html}
}

@inproceedings{dao2022flashattention,
  address   = {New Orleans, LA, USA},
  author    = {Tri Dao and
               Daniel Y. Fu and
               Stefano Ermon and
               Atri Rudra and
               Christopher R{\'{e}}},
  title     = {{FlashAttention}: Fast and Memory-Efficient Exact Attention with {IO}-Awareness},
  booktitle = {The Thirty-Sixth Annual Conference on Neural Information Processing Systems},
  year      = {2022},
  url       = {http://papers.nips.cc/paper\_files/paper/2022/hash/67d57c32e20fd0a7a302cb81d36e40d5-Abstract-Conference.html}
}

@misc{chatgpt,
  author       = {OpenAI},
  howpublished = {\url{https://chat.chatbotapp.ai/}},
  title        = {ChatGPT},
  year         = {2025},
  note         = {Accessed: 2025-08-01}
}

@misc{claude,
  author       = {Anthropic},
  howpublished = {\url{https://www.anthropic.com/claude}},
  title        = {Claude},
  year         = {2025},
  note         = {Accessed: 2025-08-01}
}

@misc{gemini,
  author       = {Google},
  howpublished = {\url{https://gemini.google.com/app}},
  title        = {Gemini},
  year         = {2025},
  note         = {Accessed: 2025-08-01}
}

@misc{qwen3-14b,
  author       = {Qwen},
  howpublished = {\url{https://huggingface.co/Qwen/Qwen3-14B}},
  title        = {Qwen3-14B},
  year         = {2025},
  note         = {Accessed: 2026-04-07}
}

@misc{qwen3-32b,
  author       = {Qwen},
  howpublished = {\url{https://huggingface.co/Qwen/Qwen3-32B}},
  title        = {Qwen3-32B},
  year         = {2025},
  note         = {Accessed: 2026-04-07}
}

@misc{qwen3,
  title         = {Qwen3 Technical Report},
  author        = {Qwen Team},
  year          = {2025},
  eprint        = {2505.09388},
  archiveprefix = {arXiv},
  primaryclass  = {cs.CL},
  url           = {https://arxiv.org/abs/2505.09388}
}

@misc{llama3.1-70B,
  author       = {Meta},
  howpublished = {\url{https://huggingface.co/meta-llama/Llama-3.1-70B}},
  title        = {Llama-3.1-70B},
  year         = {2024},
  note         = {Accessed: 2024-09-25}
}

@misc{llama4,
  author       = {Meta},
  howpublished = {\url{https://ai.meta.com/blog/llama-4-multimodal-intelligence/}},
  title        = {The Llama 4 herd: The beginning of a new era of natively multimodal AI innovation},
  year         = {2025},
  note         = {Accessed: 2025-04-05}
}

@article{liu2024retrievalattention,
  author  = {Di Liu and
             Meng Chen and
             Baotong Lu and
             Huiqiang Jiang and
             Zhenhua Han and
             Qianxi Zhang and
             Qi Chen and
             Chengruidong Zhang and
             Bailu Ding and
             Kai Zhang and
             Chen Chen and
             Fan Yang and
             Yuqing Yang and
             Lili Qiu},
  title   = {{RetrievalAttention}: Accelerating Long-Context {LLM} Inference via Vector Retrieval},
  journal = {CoRR},
  volume  = {abs/2409.10516},
  year    = {2024},
  doi     = {10.48550/ARXIV.2409.10516}
}

@inproceedings{streamingllm,
  address   = {Vienna, Austria},
  author    = {Guangxuan Xiao and
               Yuandong Tian and
               Beidi Chen and
               Song Han and
               Mike Lewis},
  title     = {Efficient Streaming Language Models with Attention Sinks},
  booktitle = {The Twelfth International Conference on Learning Representations},
  publisher = {OpenReview.net},
  year      = {2024},
  url       = {https://openreview.net/forum?id=NG7sS51zVF}
}

@inproceedings{h2o,
  address   = {New Orleans, LA, USA},
  author    = {Zhenyu Zhang and
               Ying Sheng and
               Tianyi Zhou and
               Tianlong Chen and
               Lianmin Zheng and
               Ruisi Cai and
               Zhao Song and
               Yuandong Tian and
               Christopher R{\'{e}} and
               Clark W. Barrett and
               Zhangyang Wang and
               Beidi Chen},
  title     = {{H2O:} Heavy-Hitter Oracle for Efficient Generative Inference of Large Language Models},
  booktitle = {The Thirty-Seventh Annual Conference on Neural Information Processing Systems},
  year      = {2023},
  url       = {http://papers.nips.cc/paper\_files/paper/2023/hash/6ceefa7b15572587b78ecfcebb2827f8-Abstract-Conference.html}
}

@inproceedings{li2024snapkv,
  address   = {Vancouver, BC, Canada},
  author    = {Yuhong Li and
               Yingbing Huang and
               Bowen Yang and
               Bharat Venkitesh and
               Acyr Locatelli and
               Hanchen Ye and
               Tianle Cai and
               Patrick Lewis and
               Deming Chen},
  title     = {SnapKV: {LLM} Knows What You are Looking for Before Generation},
  booktitle = {The Thirty-Eighth Annual Conference on Neural Information Processing Systems},
  year      = {2024},
  url       = {http://papers.nips.cc/paper\_files/paper/2024/hash/28ab418242603e0f7323e54185d19bde-Abstract-Conference.html}
}

@inproceedings{magicpig,
  address   = {Singapore},
  author    = {Zhuoming Chen and
               Ranajoy Sadhukhan and
               Zihao Ye and
               Yang Zhou and
               Jianyu Zhang and
               Niklas Nolte and
               Yuandong Tian and
               Matthijs Douze and
               L{\'{e}}on Bottou and
               Zhihao Jia and
               Beidi Chen},
  title     = {{MagicPIG}: {LSH} Sampling for Efficient {LLM} Generation},
  booktitle = {The Thirteenth International Conference on Learning Representations},
  publisher = {OpenReview.net},
  year      = {2025},
  url       = {https://openreview.net/forum?id=ALzTQUgW8a}
}

@article{child2019generating,
  author  = {Rewon Child and
             Scott Gray and
             Alec Radford and
             Ilya Sutskever},
  title   = {Generating Long Sequences with Sparse Transformers},
  journal = {CoRR},
  volume  = {abs/1904.10509},
  year    = {2019},
  url     = {http://arxiv.org/abs/1904.10509}
}

@inproceedings{ribarsparq,
  address   = {Vienna, Austria},
  author    = {Luka Ribar and
               Ivan Chelombiev and
               Luke Hudlass{-}Galley and
               Charlie Blake and
               Carlo Luschi and
               Douglas Orr},
  title     = {{SparQ} Attention: Bandwidth-Efficient {LLM} Inference},
  booktitle = {Forty-first International Conference on Machine Learning},
  publisher = {OpenReview.net},
  year      = {2024},
  url       = {https://openreview.net/forum?id=OS5dqxmmtl}
}

@inproceedings{lee2024InfiniGen,
  address   = {Santa Clara, CA},
  author    = {Wonbeom Lee and Jungi Lee and Junghwan Seo and Jaewoong Sim},
  booktitle = {18th USENIX Symposium on Operating Systems Design and Implementation},
  pages     = {155--172},
  publisher = {{USENIX} Association},
  title     = {{InfiniGen}: Efficient Generative Inference of Large Language Models with Dynamic {KV} Cache Management},
  url       = {https://www.usenix.org/conference/osdi24/presentation/lee},
  year      = {2024}
}

@inproceedings{zheng2024sglang,
  address   = {Vancouver, BC, Canada},
  author    = {Lianmin Zheng and
               Liangsheng Yin and
               Zhiqiang Xie and
               Chuyue Sun and
               Jeff Huang and
               Cody Hao Yu and
               Shiyi Cao and
               Christos Kozyrakis and
               Ion Stoica and
               Joseph E. Gonzalez and
               Clark W. Barrett and
               Ying Sheng},
  title     = {{SGLang}: Efficient Execution of Structured Language Model Programs},
  booktitle = {The Thirty-Eighth Annual Conference on Neural Information Processing Systems},
  year      = {2024},
  url       = {http://papers.nips.cc/paper\_files/paper/2024/hash/724be4472168f31ba1c9ac630f15dec8-Abstract-Conference.html}
}

@article{bairi2024codeplan,
  author  = {Ramakrishna Bairi and
             Atharv Sonwane and
             Aditya Kanade and
             Vageesh D. C. and
             Arun Iyer and
             Suresh Parthasarathy and
             Sriram K. Rajamani and
             Balasubramanyan Ashok and
             Shashank Shet},
  title   = {{CodePlan}: Repository-Level Coding using LLMs and Planning},
  journal = {Proceedings of the ACM on Software Engineering},
  volume  = {1},
  number  = {{FSE}},
  pages   = {675--698},
  year    = {2024},
  doi     = {10.1145/3643757}
}

@inproceedings{qin2025mooncake,
  address   = {Santa Clara, CA, USA},
  author    = {Ruoyu Qin and
               Zheming Li and
               Weiran He and
               Jialei Cui and
               Feng Ren and
               Mingxing Zhang and
               Yongwei Wu and
               Weimin Zheng and
               Xinran Xu},
  title     = {{Mooncake}: Trading More Storage for Less Computation - {A} KVCache-centric Architecture for Serving {LLM} Chatbot},
  booktitle = {23rd {USENIX} Conference on File and Storage Technologies},
  pages     = {155--170},
  publisher = {{USENIX} Association},
  year      = {2025},
  url       = {https://www.usenix.org/conference/fast25/presentation/qin}
}

@inproceedings{lin2024awq,
  address   = {Santa Clara, CA, USA},
  author    = {Ji Lin and
               Jiaming Tang and
               Haotian Tang and
               Shang Yang and
               Wei{-}Ming Chen and
               Wei{-}Chen Wang and
               Guangxuan Xiao and
               Xingyu Dang and
               Chuang Gan and
               Song Han},
  title     = {{AWQ:} Activation-aware Weight Quantization for On-Device {LLM} Compression and Acceleration},
  booktitle = {Proceedings of the Seventh Annual Conference on Machine Learning and Systems},
  publisher = {mlsys.org},
  year      = {2024},
  url       = {https://proceedings.mlsys.org/paper\_files/paper/2024/hash/42a452cbafa9dd64e9ba4aa95cc1ef21-Abstract-Conference.html}
}

@inproceedings{zhao2024atom,
  address   = {Santa Clara, CA, USA},
  author    = {Yilong Zhao and
               Chien{-}Yu Lin and
               Kan Zhu and
               Zihao Ye and
               Lequn Chen and
               Size Zheng and
               Luis Ceze and
               Arvind Krishnamurthy and
               Tianqi Chen and
               Baris Kasikci},
  title     = {Atom: Low-Bit Quantization for Efficient and Accurate {LLM} Serving},
  booktitle = {Proceedings of the Seventh Annual Conference on Machine Learning and Systems},
  publisher = {mlsys.org},
  year      = {2024},
  url       = {https://proceedings.mlsys.org/paper\_files/paper/2024/hash/5edb57c05c81d04beb716ef1d542fe9e-Abstract-Conference.html}
}

@inproceedings{alizadeh2024llm,
  address   = {Bangkok, Thailand},
  author    = {Keivan Alizadeh and
               Iman Mirzadeh and
               Dmitry Belenko and
               S. Khatamifard and
               Minsik Cho and
               Carlo C. del Mundo and
               Mohammad Rastegari and
               Mehrdad Farajtabar},
  title     = {{LLM} in a flash: Efficient Large Language Model Inference with Limited Memory},
  booktitle = {Proceedings of the 62nd Annual Meeting of the Association for Computational Linguistics (Volume 1: Long Papers)},
  pages     = {12562--12584},
  publisher = {Association for Computational Linguistics},
  year      = {2024},
  doi       = {10.18653/V1/2024.ACL-LONG.678}
}

@inproceedings{song2024powerinfer,
  address   = {Austin, TX, USA},
  author    = {Yixin Song and
               Zeyu Mi and
               Haotong Xie and
               Haibo Chen},
  title     = {{PowerInfer}: Fast Large Language Model Serving with a Consumer-grade {GPU}},
  booktitle = {Proceedings of the {ACM} {SIGOPS} 30th Symposium on Operating Systems Principles},
  pages     = {590--606},
  publisher = {{ACM}},
  year      = {2024},
  doi       = {10.1145/3694715.3695964}
}

@article{xue2024powerinfer,
  author  = {Zhenliang Xue and
             Yixin Song and
             Zeyu Mi and
             Le Chen and
             Yubin Xia and
             Haibo Chen},
  title   = {{PowerInfer-2}: Fast Large Language Model Inference on a Smartphone},
  journal = {CoRR},
  volume  = {abs/2406.06282},
  year    = {2024},
  doi     = {10.48550/ARXIV.2406.06282}
}

@inproceedings{jiang2024minference,
  address   = {Vancouver, BC, Canada},
  author    = {Huiqiang Jiang and
               Yucheng Li and
               Chengruidong Zhang and
               Qianhui Wu and
               Xufang Luo and
               Surin Ahn and
               Zhenhua Han and
               Amir H. Abdi and
               Dongsheng Li and
               Chin{-}Yew Lin and
               Yuqing Yang and
               Lili Qiu},
  title     = {MInference 1.0: Accelerating Pre-filling for Long-Context LLMs via Dynamic Sparse Attention},
  booktitle = {The Thirty-Eighth Annual Conference on Neural Information Processing Systems},
  year      = {2024},
  url       = {http://papers.nips.cc/paper\_files/paper/2024/hash/5dfbe6f5671e82c76841ba687a8a9ecb-Abstract-Conference.html}
}

@article{liu2025deepseek,
  title   = {Deepseek-v3. 2: Pushing the frontier of open large language models},
  author  = {Liu, Aixin and Mei, Aoxue and Lin, Bangcai and Xue, Bing and Wang, Bingxuan and Xu, Bingzheng and Wu, Bochao and Zhang, Bowei and Lin, Chaofan and Dong, Chen and others},
  journal = {arXiv preprint arXiv:2512.02556},
  year    = {2025}
}

@inproceedings{yuan2025native,
  address   = {Vienna, Austria},
  author    = {Jingyang Yuan and
               Huazuo Gao and
               Damai Dai and
               Junyu Luo and
               Liang Zhao and
               Zhengyan Zhang and
               Zhenda Xie and
               Yuxing Wei and
               Lean Wang and
               Zhiping Xiao and
               Yuqing Wang and
               Chong Ruan and
               Ming Zhang and
               Wenfeng Liang and
               Wangding Zeng},
  title     = {Native Sparse Attention: Hardware-Aligned and Natively Trainable Sparse Attention},
  booktitle = {Proceedings of the 63rd Annual Meeting of the Association for Computational Linguistics (Volume 1: Long Papers)},
  pages     = {23078--23097},
  publisher = {Association for Computational Linguistics},
  year      = {2025},
  url       = {https://aclanthology.org/2025.acl-long.1126/}
}

@inproceedings{yu2022orca,
  address   = {Carlsbad, CA, USA},
  author    = {Gyeong{-}In Yu and
               Joo Seong Jeong and
               Geon{-}Woo Kim and
               Soojeong Kim and
               Byung{-}Gon Chun},
  title     = {Orca: {A} Distributed Serving System for Transformer-Based Generative Models},
  booktitle = {16th {USENIX} Symposium on Operating Systems Design and Implementation},
  pages     = {521--538},
  publisher = {{USENIX} Association},
  year      = {2022},
  url       = {https://www.usenix.org/conference/osdi22/presentation/yu}
}

@inproceedings{kwon2023efficient,
  address   = {Koblenz, Germany},
  author    = {Woosuk Kwon and
               Zhuohan Li and
               Siyuan Zhuang and
               Ying Sheng and
               Lianmin Zheng and
               Cody Hao Yu and
               Joseph Gonzalez and
               Hao Zhang and
               Ion Stoica},
  title     = {Efficient Memory Management for Large Language Model Serving with PagedAttention},
  booktitle = {Proceedings of the 29th Symposium on Operating Systems Principles},
  pages     = {611--626},
  publisher = {{ACM}},
  year      = {2023},
  doi       = {10.1145/3600006.3613165}
}

@inproceedings{pope2023efficiently,
  address   = {Miami, FL, USA},
  author    = {Reiner Pope and
               Sholto Douglas and
               Aakanksha Chowdhery and
               Jacob Devlin and
               James Bradbury and
               Jonathan Heek and
               Kefan Xiao and
               Shivani Agrawal and
               Jeff Dean},
  title     = {Efficiently Scaling Transformer Inference},
  booktitle = {Proceedings of the Sixth Conference on Machine Learning and Systems},
  publisher = {mlsys.org},
  year      = {2023},
  url       = {https://proceedings.mlsys.org/paper\_files/paper/2023/hash/c4be71ab8d24cdfb45e3d06dbfca2780-Abstract-mlsys2023.html}
}

@inproceedings{wu2024loongserve,
  address   = {Austin, TX, USA},
  author    = {Bingyang Wu and
               Shengyu Liu and
               Yinmin Zhong and
               Peng Sun and
               Xuanzhe Liu and
               Xin Jin},
  title     = {{LoongServe}: Efficiently Serving Long-Context Large Language Models with Elastic Sequence Parallelism},
  booktitle = {Proceedings of the {ACM} {SIGOPS} 30th Symposium on Operating Systems Principles},
  pages     = {640--654},
  publisher = {{ACM}},
  year      = {2024},
  doi       = {10.1145/3694715.3695948}
}

@inproceedings{zhong2024distserve,
  address   = {Santa Clara, CA, USA},
  author    = {Yinmin Zhong and
               Shengyu Liu and
               Junda Chen and
               Jianbo Hu and
               Yibo Zhu and
               Xuanzhe Liu and
               Xin Jin and
               Hao Zhang},
  title     = {{DistServe}: Disaggregating Prefill and Decoding for Goodput-optimized Large Language Model Serving},
  booktitle = {18th {USENIX} Symposium on Operating Systems Design and Implementation},
  pages     = {193--210},
  publisher = {{USENIX} Association},
  year      = {2024},
  url       = {https://www.usenix.org/conference/osdi24/presentation/zhong-yinmin}
}

@inproceedings{agrawal2024taming,
  address   = {Santa Clara, CA, USA},
  author    = {Amey Agrawal and
               Nitin Kedia and
               Ashish Panwar and
               Jayashree Mohan and
               Nipun Kwatra and
               Bhargav S. Gulavani and
               Alexey Tumanov and
               Ramachandran Ramjee},
  title     = {Taming Throughput-Latency Tradeoff in {LLM} Inference with Sarathi-Serve},
  booktitle = {18th {USENIX} Symposium on Operating Systems Design and Implementation},
  pages     = {117--134},
  publisher = {{USENIX} Association},
  year      = {2024},
  url       = {https://www.usenix.org/conference/osdi24/presentation/agrawal}
}

@article{xu2024recycled,
  author  = {Fangyuan Xu and
             Tanya Goyal and
             Eunsol Choi},
  title   = {Recycled Attention: Efficient inference for long-context language models},
  journal = {CoRR},
  volume  = {abs/2411.05787},
  year    = {2024},
  doi     = {10.48550/ARXIV.2411.05787}
}

@inproceedings{xiao2024duoattention,
  address   = {Singapore},
  author    = {Guangxuan Xiao and
               Jiaming Tang and
               Jingwei Zuo and
               Junxian Guo and
               Shang Yang and
               Haotian Tang and
               Yao Fu and
               Song Han},
  title     = {{DuoAttention}: Efficient Long-Context {LLM} Inference with Retrieval and Streaming Heads},
  booktitle = {The Thirteenth International Conference on Learning Representations},
  publisher = {OpenReview.net},
  year      = {2025},
  url       = {https://openreview.net/forum?id=cFu7ze7xUm}
}

@article{yang2025lserve,
  author  = {Shang Yang and
             Junxian Guo and
             Haotian Tang and
             Qinghao Hu and
             Guangxuan Xiao and
             Jiaming Tang and
             Yujun Lin and
             Zhijian Liu and
             Yao Lu and
             Song Han},
  title   = {{LServe}: Efficient Long-sequence {LLM} Serving with Unified Sparse Attention},
  journal = {CoRR},
  volume  = {abs/2502.14866},
  year    = {2025},
  doi     = {10.48550/ARXIV.2502.14866}
}

@inproceedings{hooper2024kvquant,
  address   = {Vancouver, BC, Canada},
  author    = {Coleman Hooper and
               Sehoon Kim and
               Hiva Mohammadzadeh and
               Michael W. Mahoney and
               Yakun Sophia Shao and
               Kurt Keutzer and
               Amir Gholami},
  title     = {{KVQuant}: Towards 10 Million Context Length {LLM} Inference with {KV} Cache Quantization},
  booktitle = {The Thirty-Eighth Annual Conference on Neural Information Processing Systems},
  year      = {2024},
  url       = {http://papers.nips.cc/paper\_files/paper/2024/hash/028fcbcf85435d39a40c4d61b42c99a4-Abstract-Conference.html}
}

@inproceedings{zhang2024kv,
  address   = {Vancouver, BC, Canada},
  author    = {Tianyi Zhang and
               Jonah Yi and
               Zhaozhuo Xu and
               Anshumali Shrivastava},
  title     = {{KV} Cache is 1 Bit Per Channel: Efficient Large Language Model Inference with Coupled Quantization},
  booktitle = {The Thirty-Eighth Annual Conference on Neural Information Processing Systems},
  year      = {2024},
  url       = {http://papers.nips.cc/paper\_files/paper/2024/hash/05d6b5b6901fb57d2c287e1d3ce6d63c-Abstract-Conference.html}
}

@article{wan2024d2o,
  author  = {Zhongwei Wan and
             Xinjian Wu and
             Yu Zhang and
             Yi Xin and
             Chaofan Tao and
             Zhihong Zhu and
             Xin Wang and
             Siqi Luo and
             Jing Xiong and
             Mi Zhang},
  title   = {{D2O:} Dynamic Discriminative Operations for Efficient Generative Inference of Large Language Models},
  journal = {CoRR},
  volume  = {abs/2406.13035},
  year    = {2024},
  doi     = {10.48550/ARXIV.2406.13035}
}

@article{gao2025seerattention-r,
  title   = {Seerattention-r: Sparse attention adaptation for long reasoning},
  author  = {Gao, Yizhao and Guo, Shuming and Cao, Shijie and Xia, Yuqing and Cheng, Yu and Wang, Lei and Ma, Lingxiao and Sun, Yutao and Ye, Tianzhu and Dong, Li and others},
  journal = {arXiv preprint arXiv:2506.08889},
  year    = {2025}
}

@article{chen2023accelerating,
  author  = {Charlie Chen and
             Sebastian Borgeaud and
             Geoffrey Irving and
             Jean{-}Baptiste Lespiau and
             Laurent Sifre and
             John Jumper},
  title   = {Accelerating Large Language Model Decoding with Speculative Sampling},
  journal = {CoRR},
  volume  = {abs/2302.01318},
  year    = {2023},
  doi     = {10.48550/ARXIV.2302.01318}
}

@inproceedings{xuxattention,
  address   = {Vancouver, BC, Canada},
  author    = {Ruyi Xu and
               Guangxuan Xiao and
               Haofeng Huang and
               Junxian Guo and
               Song Han},
  title     = {{XAttention}: Block Sparse Attention with Antidiagonal Scoring},
  booktitle = {Forty-second International Conference on Machine Learning},
  publisher = {OpenReview.net},
  year      = {2025},
  url       = {https://openreview.net/forum?id=KG6aBfGi6e}
}

@article{agrawal2023sarathi,
  author  = {Amey Agrawal and
             Ashish Panwar and
             Jayashree Mohan and
             Nipun Kwatra and
             Bhargav S. Gulavani and
             Ramachandran Ramjee},
  title   = {{SARATHI:} Efficient {LLM} Inference by Piggybacking Decodes with Chunked Prefills},
  journal = {CoRR},
  volume  = {abs/2308.16369},
  year    = {2023},
  doi     = {10.48550/ARXIV.2308.16369}
}

@article{jin2024long,
  title   = {Long-context llms meet rag: Overcoming challenges for long inputs in rag},
  author  = {Jin, Bowen and Yoon, Jinsung and Han, Jiawei and Arik, Sercan O},
  journal = {arXiv preprint arXiv:2410.05983},
  year    = {2024}
}

@inproceedings{wei2022chain,
  address   = {New Orleans, LA, USA},
  author    = {Jason Wei and
               Xuezhi Wang and
               Dale Schuurmans and
               Maarten Bosma and
               Brian Ichter and
               Fei Xia and
               Ed H. Chi and
               Quoc V. Le and
               Denny Zhou},
  title     = {Chain-of-Thought Prompting Elicits Reasoning in Large Language Models},
  booktitle = {The Thirty-Sixth Annual Conference on Neural Information Processing Systems},
  year      = {2022},
  url       = {http://papers.nips.cc/paper\_files/paper/2022/hash/9d5609613524ecf4f15af0f7b31abca4-Abstract-Conference.html}
}

@inproceedings{correia2019adaptively,
  address   = {Hong Kong, China},
  author    = {Gon{\c{c}}alo M. Correia and
               Vlad Niculae and
               Andr{\'{e}} F. T. Martins},
  title     = {Adaptively Sparse Transformers},
  booktitle = {Proceedings of the 2019 Conference on Empirical Methods in Natural Language Processing and the 9th International Joint Conference on Natural Language Processing},
  pages     = {2174--2184},
  publisher = {Association for Computational Linguistics},
  year      = {2019},
  doi       = {10.18653/V1/D19-1223}
}

@article{deng2024sparse,
  title   = {How Sparse Attention Approximates Exact Attention? Your Attention is Naturally $n^{C}$-Sparse},
  author  = {Yichuan Deng and 
             Zhao Song and 
             Jing Xiong and 
             Chiwun Yang},
  journal = {arXiv preprint arXiv:2404.02690},
  year    = {2024}
}

@article{longbench-v2,
  title   = {LongBench v2: Towards Deeper Understanding and Reasoning on Realistic Long-context Multitasks},
  author  = {Yushi Bai and Shangqing Tu and Jiajie Zhang and Hao Peng and Xiaozhi Wang and Xin Lv and Shulin Cao and Jiazheng Xu and Lei Hou and Yuxiao Dong and Jie Tang and Juanzi Li},
  journal = {arXiv preprint arXiv:2412.15204},
  year    = {2024}
}

@misc{liu2024longgenbench,
  title         = {LongGenBench: Long-context Generation Benchmark},
  author        = {Xiang Liu and Peijie Dong and Xuming Hu and Xiaowen Chu},
  year          = {2024},
  eprint        = {2410.04199},
  archiveprefix = {arXiv},
  primaryclass  = {cs.CL},
  url           = {https://arxiv.org/abs/2410.04199}
}

@misc{ai_dynamo_aiperf_github,
  author       = {{ai-dynamo}},
  title        = {{AIPerf}: A comprehensive benchmarking tool that measures the performance of generative AI models served by your preferred inference solution},
  howpublished = {\url{https://github.com/ai-dynamo/aiperf}},
  note         = {GitHub repository, accessed 2026-03-29},
  year         = {2026}
}

@inproceedings{bai-etal-2024-longbench,
    title = "{L}ong{B}ench: A Bilingual, Multitask Benchmark for Long Context Understanding",
    author = "Bai, Yushi  and
      Lv, Xin  and
      Zhang, Jiajie  and
      Lyu, Hongchang  and
      Tang, Jiankai  and
      Huang, Zhidian  and
      Du, Zhengxiao  and
      Liu, Xiao  and
      Zeng, Aohan  and
      Hou, Lei  and
      Dong, Yuxiao  and
      Tang, Jie  and
      Li, Juanzi",
    editor = "Ku, Lun-Wei  and
      Martins, Andre  and
      Srikumar, Vivek",
    booktitle = "Proceedings of the 62nd Annual Meeting of the Association for Computational Linguistics (Volume 1: Long Papers)",
    month = aug,
    year = "2024",
    address = "Bangkok, Thailand",
    publisher = "Association for Computational Linguistics",
    url = "https://aclanthology.org/2024.acl-long.172/",
    doi = "10.18653/v1/2024.acl-long.172",
    pages = "3119--3137",
}

@misc{geminiteam2024gemini15unlockingmultimodal,
      title={Gemini 1.5: Unlocking multimodal understanding across millions of tokens of context}, 
      author={Gemini Team},
      year={2024},
      eprint={2403.05530},
      archivePrefix={arXiv},
      primaryClass={cs.CL},
      url={https://arxiv.org/abs/2403.05530}, 
}

@misc{wang2023docllmlayoutawaregenerativelanguage,
      title={DocLLM: A layout-aware generative language model for multimodal document understanding}, 
      author={Dongsheng Wang and Natraj Raman and Mathieu Sibue and Zhiqiang Ma and Petr Babkin and Simerjot Kaur and Yulong Pei and Armineh Nourbakhsh and Xiaomo Liu},
      year={2023},
      eprint={2401.00908},
      archivePrefix={arXiv},
      primaryClass={cs.CL},
      url={https://arxiv.org/abs/2401.00908}, 
}

@misc{ma2024mmlongbenchdocbenchmarkinglongcontextdocument,
      title={MMLongBench-Doc: Benchmarking Long-context Document Understanding with Visualizations}, 
      author={Yubo Ma and Yuhang Zang and Liangyu Chen and Meiqi Chen and Yizhu Jiao and Xinze Li and Xinyuan Lu and Ziyu Liu and Yan Ma and Xiaoyi Dong and Pan Zhang and Liangming Pan and Yu-Gang Jiang and Jiaqi Wang and Yixin Cao and Aixin Sun},
      year={2024},
      eprint={2407.01523},
      archivePrefix={arXiv},
      primaryClass={cs.CV},
      url={https://arxiv.org/abs/2407.01523}, 
}

@misc{roziere2024codellamaopenfoundation,
      title={Code Llama: Open Foundation Models for Code}, 
      author={Baptiste Rozière and Jonas Gehring and Fabian Gloeckle and Sten Sootla and Itai Gat and Xiaoqing Ellen Tan and Yossi Adi and Jingyu Liu and Romain Sauvestre and Tal Remez and Jérémy Rapin and Artyom Kozhevnikov and Ivan Evtimov and Joanna Bitton and Manish Bhatt and Cristian Canton Ferrer and Aaron Grattafiori and Wenhan Xiong and Alexandre Défossez and Jade Copet and Faisal Azhar and Hugo Touvron and Louis Martin and Nicolas Usunier and Thomas Scialom and Gabriel Synnaeve},
      year={2024},
      eprint={2308.12950},
      archivePrefix={arXiv},
      primaryClass={cs.CL},
      url={https://arxiv.org/abs/2308.12950}, 
}

@inproceedings{zhang-etal-2023-repocoder,
    title = "{R}epo{C}oder: Repository-Level Code Completion Through Iterative Retrieval and Generation",
    author = "Zhang, Fengji  and
      Chen, Bei  and
      Zhang, Yue  and
      Keung, Jacky  and
      Liu, Jin  and
      Zan, Daoguang  and
      Mao, Yi  and
      Lou, Jian-Guang  and
      Chen, Weizhu",
    editor = "Bouamor, Houda  and
      Pino, Juan  and
      Bali, Kalika",
    booktitle = "Proceedings of the 2023 Conference on Empirical Methods in Natural Language Processing",
    month = dec,
    year = "2023",
    address = "Singapore",
    publisher = "Association for Computational Linguistics",
    url = "https://aclanthology.org/2023.emnlp-main.151/",
    doi = "10.18653/v1/2023.emnlp-main.151",
    pages = "2471--2484"
}

\end{document}